\documentclass{article} 
\usepackage{iclr2027_conference,times}

\usepackage{amsmath,amsfonts,bm}

\def\eqref#1{equation~\ref{#1}}

\def\1{\bm{1}}

\DeclareMathAlphabet{\mathsfit}{\encodingdefault}{\sfdefault}{m}{sl}
\SetMathAlphabet{\mathsfit}{bold}{\encodingdefault}{\sfdefault}{bx}{n}

\usepackage{xcolor}
\usepackage{tikz}
\usetikzlibrary{fit,arrows.meta}
\usepackage{pgfplots}
\pgfplotsset{compat=1.18}
\usepackage{algorithm}
\usepackage[noend]{algpseudocode}

\usepackage[hyperfootnotes=false]{hyperref}
\usepackage{url}
\usepackage{booktabs}
\usepackage[tableposition=top]{caption}
\usepackage{graphicx}
\usepackage{siunitx}

\newif\ifpreprint
\preprinttrue

\title{Retiming Robot Demonstrations \\ with a Learned Speed Prior}

\author{Seungyong Moon\thanks{Equal contribution.} \enspace Deokjae Lee\footnotemark[1] \enspace Hansang Lee\footnotemark[1] \enspace Munkhtselmeg Battsengel \enspace Youngin Kim \\
\bf Seonghyeon Yun \enspace Sungjin Jeon \enspace SongEun Kim \enspace Jeongmin Park \enspace Hyun Oh Song\thanks{Corresponding author.} \\
Seoul National University}

\ifpreprint\iclrfinalcopy\fi
\begin{document}

\maketitle
\ifpreprint
\lhead{Preprint}
\fi

\begin{abstract}
The pace of a robot demonstration is often determined by the teleoperator rather than by the task itself. As a result, the same motion can appear at widely different speeds across demonstrations, and policies trained on such data may inherit operator-specific timing along with task behavior. We propose to learn a speed prior directly from large-scale demonstration data. Given a task instruction, a motion path and the images along it, it predicts a distribution over execution speed along the path, from which a chosen quantile defines a consistent clock for retiming each demonstration. Existing methods derive the acceleration signal from the very data they accelerate, so each new task costs another fit, annotation, or pair of predefined rates. A single trained speed prior instead covers every task in a corpus. We fit a speed prior on both robot teleoperation data and handheld demonstrations in which a human arm, rather than a robot arm, moves the gripper, and use it to retime robot demonstrations to the handheld pace before fine-tuning a vision-language-action policy. For execution, we pair the policy with a real-time optimal-control layer that tracks predicted action chunks under robot constraints using the predicted pace. On 120 RoboLab tasks, policies fine-tuned on our learned clocks succeed more often and move with up to half the jerk compared with policies fine-tuned on the original demonstration timing and uniformly accelerated to the same speed. On a real robot, our policy stacks cups in \qty{3.50}{s}, 1.72$\times$ faster than the default policy's \qty{6.02}{s} while succeeding as often, in 29 of 30 trials. Videos and further material are available at 
\ifpreprint\url{https://mllab.snu.ac.kr/projects/learned_speed_prior}\else\url{https://anonymous.4open.science/w/anony_repo-882B/}\fi.
\end{abstract}

\section{Introduction}

The pace of robot demonstrations depends on both the collection setup and the operator. Moving a robot through a teleoperation interface is slower than moving a handheld gripper directly~\citep{chi2024universal}, and novice operators often hesitate or pause where experienced operators do not. Policies trained to imitate these demonstrations inherit the operators' pace along with the task behavior, often executing more slowly than necessary~\citep{guo2025demospeedup,arachchige2025sail}.

Prior work speeds up such policies by retiming their demonstrations before or during training, or by accelerating their execution~\citep{guo2025demospeedup,kim2026espada,sanchez2026volt,nam2025speedaug,jing2026tempovla,hu2026autospeed,arachchige2025sail,kim2026time}. These methods shorten motion segments that require less precision, train the policy at a range of speeds, or execute the motion as fast as the robot's limits allow. They typically adjust the original demonstration timing based on task requirements, external speed commands, or robot limits. Few instead learn the pace from demonstrators who perform similar motions faster.

We instead draw the pace from such demonstrators, for example those using a handheld gripper or experienced with the task, and give it to the demonstrations that the policy learns from. Their demonstrations need not follow the same paths, so we do not copy their timing but learn a speed prior, a model of how fast they move along a given path. Given the local shape of an end-effector path, the wrist-camera images along it and the task instruction, the prior predicts a distribution over the speed at which demonstrators traverse each part of the path. It is trained once on both kinds of demonstrations across all their tasks, with a label that marks the kind of each demonstration, so reading a path with the label of the faster demonstrators predicts how fast they would traverse it. A quantile of this distribution defines a clock that keeps the path of a demonstration but replaces its timing. We then fine-tune a policy on demonstrations retimed with the median clock and let it also predict tempo channels, the pace of higher quantiles, so that it can be sped up at deployment.

The predicted pace reflects human motion and may therefore exceed the velocity, acceleration or torque limits of the robot. Joint PD control, commonly used to execute such policies, does not plan within these limits, so fast targets can cause tracking lag or safety reflexes~\citep{khazatsky2024droid,intelligence2025openpi}. Consecutive action chunks may also disagree at their boundaries, and each new chunk must accommodate the pace of the ongoing motion. We therefore introduce a controller that treats the predicted pace as an upper bound and executes each chunk as fast as the robot's limits permit, slowing down only where a limit becomes active. When a new chunk arrives, the controller smoothly connects it to the ongoing motion, preserving continuity in position, velocity, and acceleration while carrying the execution clock forward.

We train the speed prior on teleoperated demonstrations from DROID and handheld demonstrations from FastUMI-100K, and use it to give DROID the handheld pace before fine-tuning the released $\pi_{0.5}$ DROID policy~\citep{khazatsky2024droid,liu2025fastumi,intelligence2025pi05,intelligence2025openpi}. On 120 tasks of the RoboLab simulation benchmark, this policy succeeds more often than a uniformly accelerated policy at the same speed-up, and at $2.4\times$ it does so with 52\% less jerk~\citep{yang2026robolab}. Training only on fast demonstrations does not achieve this. A policy trained on the fastest DROID episodes alone, although their pace is $1.5\times$ that of all episodes, runs just $1.07\times$ faster than the policy trained on all of them. We observe the same on a real robot, where we adapt the prior to handheld demonstrations collected with a YUBI gripper and retime them to the pace of the shortest quarter of the episodes~\citep{ohkawa2026yubi}. Our policy stacks cups in \qty{3.50}{s} instead of \qty{6.02}{s} and succeeds in 29 of 30 trials, whereas a policy trained only on this quarter succeeds in 9 of 30 trials. On the same three-cup pyramid, SAIL reports \qty{14.71}{s} at a success rate of 0.40 for a diffusion policy trained on teleoperated demonstrations and accelerated by $1.76\times$~\citep{arachchige2025sail}, so our policy stacks them $4.2\times$ faster and succeeds more than twice as often. Finally, executing a human pace requires a controller that respects the robot's limits. With our controller in place of joint PD control, even the policy trained on the recorded timing folds the towel in 97\% of trials instead of 13\%.

\section{Related work}

\paragraph{Retiming demonstrations.} A line of work changes the timing of demonstrations before policy training. Arc-length warping reparameterizes each trajectory by path length and leaves the speed to be chosen afterward~\citep{braglia2025arc}. ISR keeps a subset of the frames, spaced at roughly equal information, measured by velocity and acceleration~\citep{yang2026improving}. Others downsample the segments that do not require precision. DemoSpeedup finds them by the action entropy of a proxy policy trained on the task~\citep{guo2025demospeedup}. ESPADA labels a seed demonstration with vision-language and language models and propagates the labels to the other demonstrations of the task~\citep{kim2026espada}. VOLT prompts a vision-language model with examples from other tasks to label every demonstration~\citep{sanchez2026volt}. In these methods, the timing follows fixed criteria or is compressed by factors chosen by the user.

\paragraph{Training policies at different speeds.} Other methods retime demonstrations to a range of speeds during training and differ in how they choose the executed speed. SpeedAug trains the policy on demonstrations accelerated at random rates and then fine-tunes it with reinforcement learning in the task environment~\citep{nam2025speedaug}. TempoVLA conditions a vision-language-action policy on a speed command, which a vision-language model chooses at deployment~\citep{jing2026tempovla}. AutoSpeed learns a per-timestep speed by selecting, among candidates retimed at a discrete set of ratios, the one that best trades the policy's own prediction error against its prediction horizon~\citep{hu2026autospeed}. In all of these methods, the new timing is derived from the demonstration itself, the task at hand or an external command. In contrast, we draw the pace from other demonstrators who perform similar motions faster, and learn it as a speed prior shared across tasks.

\paragraph{Executing policies faster.} Learned policies usually send each predicted action as a position target to a PD controller, one action per time step of the demonstrations, which limits how much faster they can run~\citep{khazatsky2024droid,intelligence2025openpi}. SAIL predicts whether the current motion is critical, using labels obtained by clustering waypoints of the demonstrations, and scales the duration of each step by one of two factors~\citep{arachchige2025sail}. Realtime-VLA V2 retimes each chunk under velocity bounds with speeds from a model trained on the speed commands an operator gives during rollouts~\citep{yang2026realtime}. RACE retimes each chunk by time-optimal path parameterization, as fast as the velocity and torque limits of the robot allow~\citep{pham2018new,kim2026time}. Consecutive chunks are reconciled inside the policy or by the executor, as in VLA-RAIL, which joins them with quintic splines but scales their timing uniformly~\citep{black2025real,black2025training,zhao2025vla}. Other policies output controller parameters, such as the stiffness of an impedance controller~\citep{martinmartin2019variable,kamijo2024learning,hou2025adaptive,kim2026compvla}. Our policy likewise outputs its pace, and our controller follows this pace within the robot's dynamic limits and joins consecutive chunks without a jump in position, velocity or acceleration.

\section{Learning the pace from demonstrations}
\label{sec:method}

\subsection{Paths and clocks}
\label{sec:overview}
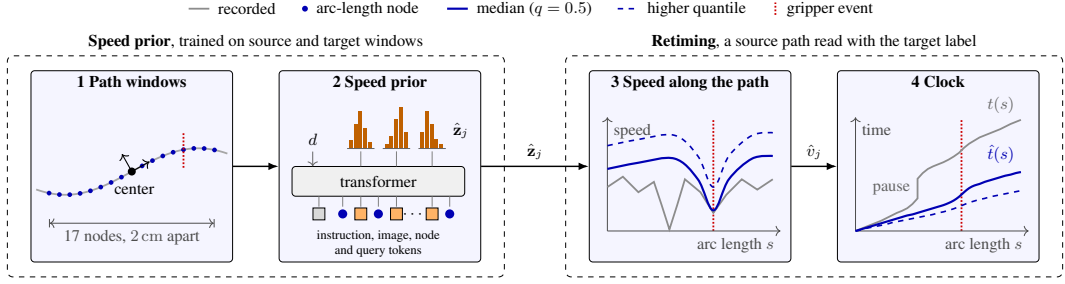
\begin{figure}[t]
    \centering
    \resizebox{\linewidth}{!}{%
    \begin{tikzpicture}[font=\small, black,
      blk/.style={draw, rounded corners=2pt, align=center, inner sep=4pt, fill=white},
      stp/.style={draw, rounded corners=2pt, fill=blue!4, minimum width=3.9cm, minimum height=3.7cm},
      arr/.style={-{Latex[length=2mm]}, line width=0.8pt},
      ax/.style={->, line width=0.5pt, black!70},
      tok/.style={draw, minimum size=0.2cm, inner sep=0pt}]
      \begin{scope}[shift={(5.3,3.05)}, font=\footnotesize]
        \draw[black!45, line width=1pt] (0,0) -- (0.4,0); \node[anchor=west] at (0.45,0) {recorded};
        \fill[blue!70!black] (2.25,0) circle (1.6pt); \node[anchor=west] at (2.35,0) {arc-length node};
        \draw[blue!70!black, line width=1.2pt] (5.0,0) -- (5.4,0); \node[anchor=west] at (5.45,0) {median ($q = 0.5$)};
        \draw[blue!70!black, dashed, line width=0.9pt] (8.3,0) -- (8.7,0); \node[anchor=west] at (8.75,0) {higher quantile};
        \draw[red!80!black, densely dotted, line width=1pt] (11.35,-0.12) -- (11.35,0.12); \node[anchor=west] at (11.45,0) {gripper event};
      \end{scope}
      \node[stp] (win) at (4.2,0) {};
      \node[stp] (pri) at (9.0,0) {};
      \node[stp] (spd) at (15.0,0) {};
      \node[stp] (clk) at (19.8,0) {};
      \foreach \n/\lab in {win/{1 Path windows}, pri/{2 Speed prior}, spd/{3 Speed along the path}, clk/{4 Clock}} {\node[anchor=north, font=\small\bfseries, inner sep=3pt] at (\n.north) {\lab};}
      \node[draw, dashed, rounded corners=3pt, fit=(win)(pri), inner xsep=12.8pt, inner ysep=8pt] (train) {};
      \node[anchor=south, align=center] at (train.north) {\textbf{Speed prior}, trained on source and target windows};
      \node[draw, dashed, rounded corners=3pt, fit=(spd)(clk), inner xsep=12.8pt, inner ysep=8pt] (ret) {};
      \node[anchor=south, align=center] at (ret.north) {\textbf{Retiming}, a source path read with the target label};
      \draw[arr] (win) -- (pri);
      \draw[arr] (pri) -- node[above, font=\footnotesize] {$\hat{\mathbf{z}}_j$} (spd);
      \draw[arr] (spd) -- node[above, font=\footnotesize] {$\hat{v}_j$} (clk);
      \begin{scope}[shift={(4.2,-0.1)}]
        \draw[black!35, line width=1pt, domain=-1.85:1.85, samples=60, smooth] plot (\x, {0.45*sin(70*\x)});
        \foreach \i in {0,...,16} {\pgfmathsetmacro{\px}{-1.6+0.2*\i} \fill[blue!70!black] (\px, {0.45*sin(70*\px)}) circle (1.3pt);}
        \draw[red!80!black, densely dotted, line width=1pt] (1.0, {0.45*sin(70)-0.35}) -- (1.0, {0.45*sin(70)+0.35});
        \fill[black] (0,0) circle (2.2pt);
        \draw[->, line width=0.6pt] (0,0) -- (0.34,0.19);
        \draw[->, line width=0.6pt] (0,0) -- (-0.19,0.34);
        \node[font=\footnotesize, anchor=north] at (0.05,-0.12) {center};
        \draw[|<->|, black!70] (-1.6,-1.0) -- node[below, font=\footnotesize, align=center] {17 nodes, \qty{2}{cm} apart} (1.6,-1.0);
      \end{scope}
      \begin{scope}[shift={(9.0,-0.1)}]
        \node[tok, minimum size=0.23cm, fill=black!15] at (-1.18,-0.82) {};
        \node[font=\footnotesize] (dom) at (-1.3,0.62) {$d$};
        \foreach \x in {-0.72,-0.03,1.35} {\fill[blue!70!black] (\x,-0.82) circle (2.5pt);}
        \foreach \x in {-0.37,0.32,1.01} {\node[tok, minimum size=0.23cm, fill=orange!60] at (\x,-0.82) {};}
        \node[font=\footnotesize] at (0.66,-0.82) {$\cdots$};
        \node[draw, rounded corners=2pt, fill=black!6, minimum width=3.34cm, minimum height=0.575cm, font=\footnotesize] (tf) at (-0.03,-0.19) {transformer};
        \foreach \x in {-1.18,-0.72,-0.37,-0.03,0.32,1.01,1.35} {\draw[black!50, line width=0.4pt] (\x,-0.68) -- (\x,-0.48);}
        \draw[->, black!60, line width=0.5pt] (dom.south) -- (-1.3,0.1);
        \foreach \x/\a/\b/\c/\d/\e in {-0.37/0.10/0.30/0.45/0.25/0.08, 0.32/0.05/0.15/0.35/0.50/0.20, 1.01/0.20/0.45/0.30/0.12/0.05} {
          \draw[black!50, line width=0.4pt] (\x,0.1) -- (\x,0.39);
          \foreach \k/\h in {0/\a, 1/\b, 2/\c, 3/\d, 4/\e} {\fill[orange!75!black] ({\x-0.23+0.092*\k}, 0.445) rectangle ++(0.075, {1.61*\h});}
          \draw[black!60, line width=0.4pt] ({\x-0.265}, 0.445) -- ({\x+0.265}, 0.445);
        }
        \node[font=\footnotesize, anchor=west] at (1.29,0.79) {$\hat{\mathbf{z}}_j$};
        \node[font=\scriptsize, align=center, anchor=north] at (-0.03,-1.02) {instruction, image, node\\and query tokens};
      \end{scope}
      \begin{scope}[shift={(15.0,-0.05)}]
        \draw[red!80!black, densely dotted, line width=1pt] (0.45,-1.2) -- (0.45,0.95);
        \draw[ax] (-1.6,-1.2) -- (1.75,-1.2) node[anchor=north east, font=\footnotesize, black!70] {arc length $s$};
        \draw[ax] (-1.6,-1.2) -- (-1.6,1.0) node[anchor=north west, font=\footnotesize, black!70] {speed};
        \draw[black!45, line width=1pt] (-1.60,-0.35) -- (-1.30,-0.15) -- (-1.00,-0.45) -- (-0.70,-0.25) -- (-0.55,-0.70) -- (-0.40,-1.17) -- (-0.25,-0.70) -- (-0.10,-0.20) -- (0.20,-0.45) -- (0.45,-0.85) -- (0.70,-0.50) -- (1.00,-0.20) -- (1.30,-0.40) -- (1.60,-0.20);
        \draw[blue!70!black, line width=1.2pt] plot[smooth, tension=0.6] coordinates {(-1.6,0) (-1,0.12) (-0.3,0.2) (0.15,-0.2) (0.45,-0.8) (0.75,-0.2) (1.2,0.2) (1.6,0.25)};
        \draw[blue!70!black, dashed, line width=0.9pt] plot[smooth, tension=0.6] coordinates {(-1.6,0.45) (-1,0.57) (-0.3,0.65) (0.15,0.25) (0.45,-0.35) (0.75,0.25) (1.2,0.65) (1.6,0.7)};
      \end{scope}
      \begin{scope}[shift={(19.8,-0.05)}]
        \draw[red!80!black, densely dotted, line width=1pt] (0.45,-1.2) -- (0.45,0.95);
        \draw[ax] (-1.6,-1.2) -- (1.75,-1.2) node[anchor=north east, font=\footnotesize, black!70] {arc length $s$};
        \draw[ax] (-1.6,-1.2) -- (-1.6,1.0) node[anchor=north west, font=\footnotesize, black!70] {time};
        \draw[black!45, line width=1pt] (-1.60,-1.20) -- (-1.50,-1.16) -- (-1.40,-1.12) -- (-1.30,-1.08) -- (-1.20,-1.04) -- (-1.10,-1.00) -- (-1.00,-0.95) -- (-0.90,-0.90) -- (-0.80,-0.85) -- (-0.70,-0.81) -- (-0.60,-0.76) -- (-0.50,-0.68) -- (-0.40,-0.57) -- (-0.40,-0.19) -- (-0.30,-0.08) -- (-0.20,0.00) -- (-0.10,0.05) -- (0.00,0.09) -- (0.10,0.13) -- (0.20,0.18) -- (0.30,0.24) -- (0.40,0.31) -- (0.50,0.41) -- (0.60,0.49) -- (0.70,0.55) -- (0.80,0.61) -- (0.90,0.65) -- (1.00,0.69) -- (1.10,0.73) -- (1.20,0.77) -- (1.30,0.82) -- (1.40,0.87) -- (1.50,0.91) -- (1.60,0.95);
        \node[font=\footnotesize, black!60, anchor=east] at (-0.42,-0.38) {pause};
        \node[font=\footnotesize, black!60, anchor=south east] at (1.6,0.95) {$t(s)$};
        \draw[blue!70!black, line width=1.2pt] plot[smooth, tension=0.5] coordinates {(-1.60,-1.20) (-1.50,-1.17) (-1.40,-1.14) (-1.30,-1.11) (-1.20,-1.08) (-1.10,-1.05) (-1.00,-1.02) (-0.90,-0.99) (-0.80,-0.96) (-0.70,-0.93) (-0.60,-0.90) (-0.50,-0.88) (-0.40,-0.85) (-0.30,-0.82) (-0.20,-0.79) (-0.10,-0.76) (0.00,-0.73) (0.10,-0.70) (0.20,-0.66) (0.30,-0.61) (0.40,-0.54) (0.50,-0.45) (0.60,-0.37) (0.70,-0.32) (0.80,-0.29) (0.90,-0.25) (1.00,-0.22) (1.10,-0.19) (1.20,-0.16) (1.30,-0.13) (1.40,-0.11) (1.50,-0.08) (1.60,-0.06)};
        \draw[blue!70!black, dashed, line width=0.9pt] plot[smooth, tension=0.5] coordinates {(-1.60,-1.20) (-1.50,-1.18) (-1.40,-1.15) (-1.30,-1.13) (-1.20,-1.11) (-1.10,-1.09) (-1.00,-1.07) (-0.90,-1.04) (-0.80,-1.02) (-0.70,-1.00) (-0.60,-0.98) (-0.50,-0.96) (-0.40,-0.94) (-0.30,-0.92) (-0.20,-0.90) (-0.10,-0.88) (0.00,-0.85) (0.10,-0.83) (0.20,-0.80) (0.30,-0.77) (0.40,-0.73) (0.50,-0.69) (0.60,-0.65) (0.70,-0.62) (0.80,-0.59) (0.90,-0.57) (1.00,-0.54) (1.10,-0.52) (1.20,-0.50) (1.30,-0.48) (1.40,-0.46) (1.50,-0.44) (1.60,-0.42)};
        \node[font=\footnotesize, blue!70!black, anchor=south east] at (1.6,-0.02) {$\hat{t}(s)$};
      \end{scope}
    \end{tikzpicture}}
    \caption{Learning the pace from demonstrations. (1) Paths are resampled every \qty{2}{cm} of arc length and read in windows of 17 nodes. (2) The prior predicts a distribution over the log speed of every interval. (3) Reading a source path with the target label gives a quantile of the speed along the path. (4) Integrating this speed gives the clock.}
    \label{fig:pace}
\end{figure}

We learn a speed prior that predicts how fast target demonstrators traverse each part of a path, and retime source demonstrations with a clock derived from it (Figure~\ref{fig:pace}). Each demonstration records end-effector positions $\mathbf{p}_k \in \mathbb{R}^3$, orientations $R_k \in SO(3)$ and gripper openings $o_k$ at timestamps $t_k$, for $k = 0, \dots, K$. We measure progress along the path with an increment that combines translation, rotation and gripper motion,
\begin{align*}
    \Delta s_k = \sqrt{\|\mathbf{p}_{k+1} - \mathbf{p}_k\|^2 + \rho_{\mathrm{r}}^2\,\theta\big(R_k^\top R_{k+1}\big)^2 + \rho_{\mathrm{g}}^2 (o_{k+1} - o_k)^2},
\end{align*}
where $\theta(\cdot)$ is the angle of a relative rotation and $\rho_{\mathrm{r}}$ and $\rho_{\mathrm{g}}$ convert radians and gripper openings to meters~\citep{park1995distance}. We set $\rho_{\mathrm{r}} = \qty{0.3}{m/rad}$, following prior work on rigid-body trajectory segmentation~\citep{verduyn2024enhancing}, and $\rho_{\mathrm{g}} = \qty{0.15}{m}$ per unit of opening. The arc length $s_k = \sum_{i<k} \Delta s_i$ grows only while the end effector or the gripper moves, so pauses do not advance it. The recorded timing is then a map $t(s)$ from arc length to time, with speed $v(s) = \mathrm{d}s/\mathrm{d}t$ along the path. Retiming replaces $t(s)$ with a clock $\hat{t}(s)$ and keeps the path.

\subsection{Training data}
\label{sec:data}

\paragraph{Source and target demonstrations.}
Source demonstrations are retimed to follow the pace that target demonstrations show. The two may come from different setups, such as teleoperation and handheld demonstrations, in which a person moves a gripper by hand~\citep{chi2024universal}. They may also come from the same setup when only some of its demonstrators move at the desired pace. Each demonstration carries a domain label $d \in \{\mathrm{src}, \mathrm{tgt}\}$. Handheld poses are expressed in the frame of the robot data, and episodes whose tracking fails are removed (Appendix~\ref{app:umi}).

\paragraph{Path windows.}
We resample each path at nodes spaced $\Delta = \qty{2}{cm}$ apart in arc length, the spacing used with this metric in prior work~\citep{verduyn2024enhancing}. The prior reads windows of 17 nodes, eight on each side of a center node, spanning \qty{32}{cm}. Each node holds its position and rotation vector in the frame of the center node, its absolute gripper opening and the nearest wrist-camera image.

\paragraph{Speed targets.}
The 17 nodes bound 16 intervals. Over interval $j$, the arm and the gripper travel $L^{\mathrm{a}}_j$ and $L^{\mathrm{g}}_j$, respectively (Appendix~\ref{app:training}). The target log speed of the interval is $y_j = \log(L_j / \Delta t_j)$, where $L_j = \max(L^{\mathrm{a}}_j, L^{\mathrm{g}}_j)$ and $\Delta t_j$ is the time between its two nodes, including any pauses.

\subsection{Speed prior}
\label{sec:prior}

\paragraph{Architecture.}
The prior is a transformer trained from scratch and conditioned on the domain label. It reads an instruction token and 17 pairs of image and node tokens interleaved with 16 query tokens. Frozen encoders embed instructions and images~\citep{zhang2025qwen3,simeoni2025dinov3}. Query tokens predict the arm speed $v^{\mathrm{a}}_j$ and the gripper speed $v^{\mathrm{g}}_j$ of their interval (Appendix~\ref{app:training}).

\paragraph{Categorical objective.}
Different demonstrators traverse similar paths at very different speeds, so we predict a distribution over the speed of each interval, not a single value, from which retiming can select a quantile. Per actuator, the prior predicts a categorical distribution over $B = 256$ equal bins of log speed from 1 to \qty{5000}{mm/s}. An interval lasts as long as its slower actuator,
\begin{align*}
  T_j = \max\big(L^{\mathrm{a}}_j / v^{\mathrm{a}}_j,\; L^{\mathrm{g}}_j / v^{\mathrm{g}}_j\big),
\end{align*}
and treating the two speeds as independent gives the distribution $\hat{\mathbf{z}}_j$ of $\log(L_j / T_j)$ on the same bins (Appendix~\ref{app:training}). The target $y_j$ is encoded as a two-hot vector $\mathbf{z}_j$~\citep{farebrother2024stop}. The prior is trained with the cross-entropy loss
\begin{align*}
  \mathcal{L} = -\frac{1}{|\mathcal{J}|} \sum_{j \in \mathcal{J}} \mathbf{z}_j^\top \log \hat{\mathbf{z}}_j,
\end{align*}
where $\mathcal{J}$ holds the intervals of the window that lie within the episode.

\subsection{Retiming}
\label{sec:retiming}

To retime a source demonstration, we ask the prior how fast a target demonstrator would traverse the same path. Since the images reveal the domain, switching the label alone does not pose this question cleanly. We therefore drop the images of a window with probability 0.5 in training. Without images, the two labels differ only in the pace of the demonstrator, and we add this difference to the source reading with images, inspired by classifier-free guidance~\citep{ho2022classifier}. Writing $f^{\mathrm{a}}(d, I)$ for the arm logits read under the label $d$ with images $I$ and $f^{\mathrm{a}}(d, \emptyset)$ for those without, the arm logits for retiming are
\begin{align*}
  \tilde{f}^{\mathrm{a}} = f^{\mathrm{a}}(\mathrm{src}, I) + \lambda\,\big[f^{\mathrm{a}}(\mathrm{tgt}, \emptyset) - f^{\mathrm{a}}(\mathrm{src}, \emptyset)\big],
\end{align*}
with a guidance coefficient $\lambda$ (Appendix~\ref{app:training}). When the two setups use different grippers, the gripper keeps the source label, since its own mechanism, not the demonstrator, sets its speed. We average the distributions $\hat{\mathbf{z}}^{(w)}_j$ that the windows $w$ covering interval $j$ predict from these logits and decode the quantile of level $q$, which gives the duration $\hat{T}_j$ and the speed $\hat{v}_j = \Delta / \hat{T}_j$. The clock assigns each recorded frame the time needed to reach its arc length at these speeds,
\begin{align*}
  \hat{t}(s) = \int_0^{s} \frac{\mathrm{d}s'}{\hat{v}(s')},
\end{align*}
where $\hat{v}(s)$ equals $\hat{v}_j$ on interval $j$. Frames at the same arc length share a timestamp, so pauses in the recording take no time on the clock.

\subsection{Policy fine-tuning}
\label{sec:policy}

\paragraph{Action targets.}
We fine-tune $\pi_{0.5}$~\citep{intelligence2025pi05} on chunks of $H$ joint-position actions from the source demonstrations retimed with the median clock ($q = 0.5$), which serves as the base clock. Observations are taken at the recorded frames, and each chunk lists the joint positions and gripper commands that the retimed demonstration reaches over the next $H$ periods of the base clock.

\paragraph{Tempo channels.}
To let the policy run faster at deployment, we also retime each demonstration with higher quantiles. The tempo channel of quantile $q$ at step $h$ is $\tau_h^{(q)} = \log\big(\delta / \delta_h^{(q)}\big)$, where the step lasts $\delta = 1/15\,\mathrm{s}$ on the base clock and $\delta_h^{(q)}$ on the clock of quantile $q$, and the policy predicts these channels as additional action dimensions (Appendix~\ref{app:policy}).

\section{Executing the predicted pace}
\label{sec:controller}

Our policy predicts actions together with a pace learned from human demonstrations (Section~\ref{sec:policy}). Executing them on a robot raises two problems. First, a human pace ignores the robot's dynamics and can exceed its velocity, acceleration or torque limits. A joint PD controller~\citep{khazatsky2024droid,intelligence2025openpi} builds its reference without regard to these limits and enforces them only reactively, by clamping and rate-limiting the torque or by the safety reflex, so fast targets make the robot lag or stop (Appendix~\ref{app:controller}). The robot should instead follow the pace wherever its limits allow and slow down only where they bind. 
Second, consecutive action chunks disagree at their boundary, which real-time chunking (RTC) reduces but does not remove~\citep{black2025real,black2025training}. Switching chunks therefore needs a transition without a jump in position, velocity or acceleration, and a timing for the new chunk. At a fixed rate, action $h$ is due $h$ control periods after its observation, but under the predicted pace the running motion runs ahead of or behind this schedule, so the new chunk should become due when the running motion reaches the same step.

\subsection{Execution cycle}
\label{sec:cycle}

The robot tracks a plan $P$, a time-parameterized reference of joint position, velocity and acceleration with gripper commands, using its own low-level controller (a \qty{1}{kHz} torque controller on our Franka Research 3). Meanwhile, each cycle runs policy inference, the solver and the installation of the new plan, which take $t_{\mathrm{vla}}$, $t_{\mathrm{solve}}$ and $t_{\mathrm{install}}$. At the observation time $t_o$, the policy maps the camera images, the joint state $\mathbf{x}_o$, the instruction and, with RTC, the unexecuted actions of $P$ to $H$ joint waypoints $\mathbf{w}_h$, gripper commands $c_h$ and tempo channels $\tau^{(q)}_h$, for the steps $h = 1, \dots, H$ after the observation. The target pace $r_h = \exp\big(\tau^{(q)}_h\big)$, or $r_h = \alpha$ for a constant speed-up $\alpha$, bounds the duration of step $h$ from below by $\delta / r_h$.

The client then reserves a switch time $t_b$ at least $t_{\mathrm{solve}} + t_{\mathrm{install}}$ ahead, so the solver plans from where $P$ will be at $t_b$ rather than from where the robot is now, and the new plan is ready before the robot gets there. It reads two quantities off $P$: its state $\mathbf{s}_b = (\mathbf{x}_b, \dot{\mathbf{x}}_b, \ddot{\mathbf{x}}_b)$ at $t_b$, and its progress $p(t)$, the number of steps that $P$ has advanced since $t_o$ on its own clock, fractional between steps. By $t_b$, $P$ has passed the steps up to $p(t_b)$, so the new chunk is kept from step $h_0 = \lfloor p(t_b) \rfloor + 1$ on, and $t^\star$ is the time at which $P$ reaches that step, $p(t^\star) = h_0$ (Appendix~\ref{app:controller} covers a $P$ that ends before). The solver receives the kept waypoints $\mathbf{w}_{h_0}, \dots, \mathbf{w}_H$, their target pace, $\mathbf{s}_b$, $t^\star$ and the robot limits, and returns a new plan $P'$. The robot switches to it at $t_b$,
\begin{align*}
  (\mathbf{x}, \dot{\mathbf{x}}, \ddot{\mathbf{x}})(t) =
  \begin{cases}
    P(t), & t < t_b, \\
    P'(t), & t \ge t_b,
  \end{cases}
  \qquad P'(t_b) = P(t_b) = \mathbf{s}_b,
\end{align*}
so $P'$ keeps the reference up to $t_b$ and replaces the rest with a transition onto the new chunk followed by the chunk itself, with no jump in position, velocity or acceleration. The policy is queried $t_{\mathrm{vla}} + t_{\mathrm{solve}} + t_{\mathrm{install}}$ before $P$ needs a successor, so the robot never waits.

\begin{figure}[t]
\centering
\resizebox{\linewidth}{!}{%
\begin{tikzpicture}[font=\small, black,
  stp/.style={draw, rounded corners=2pt, fill=blue!4, minimum width=3.9cm, minimum height=3.7cm, anchor=south west, inner sep=0pt},
  arr/.style={-{Latex[length=2mm]}, line width=0.8pt},
  run/.style={black!45, line width=1pt},
  runf/.style={black!22, line width=0.8pt},
  old/.style={circle, draw=black!50, fill=white, inner sep=0pt, minimum size=3.6pt},
  wp/.style={circle, fill=blue!70!black, inner sep=0pt, minimum size=3.6pt},
  wps/.style={circle, fill=blue!70!black, draw=white, line width=0.4pt, inner sep=0pt, minimum size=4.2pt},
  kns/.style={circle, fill=red!80!black, draw=white, line width=0.4pt, inner sep=0pt, minimum size=4.0pt},
  wpf/.style={circle, fill=blue!30, inner sep=0pt, minimum size=3.6pt},
  kn/.style={circle, fill=red!80!black, inner sep=0pt, minimum size=3.2pt},
  drf/.style={blue!70!black, line width=1.2pt},
  drff/.style={blue!35, line width=0.8pt},
  tra/.style={red!80!black, line width=1.2pt},
  traf/.style={red!35, line width=0.8pt},
  new/.style={black, line width=1.4pt},
  lbl/.style={inner sep=1pt, font=\footnotesize, black},
  sub/.style={anchor=north, inner sep=0pt, align=center, font=\footnotesize, black!60},
  box/.style={draw=blue!70!black, line width=0.4pt, minimum width=0.203cm, minimum height=0.182cm, inner sep=0pt},
  boxp/.style={draw=blue!35, line width=0.4pt, minimum width=0.203cm, minimum height=0.182cm, inner sep=0pt},
  boxf/.style={draw=blue!70!black, line width=0.4pt, minimum width=0.203cm, minimum height=0.182cm, inner sep=0pt},
  hd/.style={anchor=north, inner sep=3pt, font=\small\bfseries}]
  \node[stp] (pol) at (0.0,0) {};
  \node[hd] at (1.95,3.7) {1 Tempo policy};
  \node[sub] at (1.95,3.24) {chunk and pace};
  \node[stp] (dra) at (5.55,0) {};
  \node[hd] at (7.5,3.7) {2 Draft};
  \node[sub] at (7.5,3.24) {min-jerk path, timed at pace};
  \node[stp] (spl) at (10.35,0) {};
  \node[hd] at (12.3,3.7) {3 Splice};
  \node[sub] at (12.3,3.24) {earliest on-time entry\\under limits};
  \node[stp] (smo) at (15.15,0) {};
  \node[hd] at (17.1,3.7) {4 Smooth};
  \node[sub] at (17.1,3.24) {min-jerk QP within limits};
  \node[draw, dashed, rounded corners=3pt, fit=(dra)(spl)(smo), inner xsep=12.8pt, inner ysep=8pt] (sol) {};
  \node[anchor=south, inner sep=3pt] at (sol.north) {\textbf{Solver}, planning ahead from a reserved switch point};
  \draw[arr] (pol.east) -- (sol.west |- pol.east);
  \path ([xshift=-13.2pt]pol.west);
  \draw[arr] (dra.east) -- (spl.west);
  \draw[arr] (spl.east) -- (smo.west);
  \begin{scope}[shift={(0.0,0)}]
    \begin{scope}[yshift=0.24cm]
    \draw[run] plot[smooth] coordinates {(0.390,0.985) (0.495,1.019) (0.599,1.051) (0.704,1.082) (0.808,1.112) (0.913,1.140) (1.017,1.166) (1.122,1.191) (1.226,1.214) (1.331,1.236) (1.435,1.256) (1.540,1.274) (1.644,1.291) (1.749,1.306) (1.853,1.320) (1.958,1.332) (2.062,1.343) (2.167,1.352) (2.271,1.359) (2.376,1.365) (2.480,1.370) (2.585,1.373) (2.689,1.374) (2.794,1.373) (2.898,1.372) (3.003,1.368) (3.107,1.363) (3.212,1.356) (3.316,1.348) (3.421,1.339)};
    \node[wp] at (1.073,1.313) {}; \node[wp] at (1.442,1.432) {}; \node[wp] at (1.645,1.684) {}; \node[wp] at (1.977,1.804) {}; \node[wp] at (2.165,2.035) {}; \node[wp] at (2.431,2.180) {}; \node[wp] at (2.606,2.390) {}; \node[wp] at (2.814,2.554) {};
    \fill (0.784,1.105) circle (2pt);
    \node[lbl, anchor=north west] at ([xshift=-6pt,yshift=-5pt]0.784,1.105) {observation};
    \node[lbl, anchor=south east, blue!70!black] at ([xshift=-3pt,yshift=2pt]2.165,2.035) {new chunk};
    \node[lbl, anchor=north east] at ([yshift=-3pt]3.404,1.340) {running plan};
    \end{scope}
    \fill[orange!75!black] (1.023,0.35) rectangle (1.123,0.539);
    \fill[orange!75!black] (1.272,0.35) rectangle (1.372,0.623);
    \fill[orange!75!black] (1.520,0.35) rectangle (1.620,0.749);
    \fill[orange!75!black] (1.769,0.35) rectangle (1.869,0.791);
    \fill[orange!75!black] (2.018,0.35) rectangle (2.118,0.749);
    \fill[orange!75!black] (2.266,0.35) rectangle (2.366,0.623);
    \fill[orange!75!black] (2.515,0.35) rectangle (2.615,0.518);
    \fill[orange!75!black] (2.764,0.35) rectangle (2.864,0.476);
    \draw[black!60, line width=0.4pt] (0.953,0.35) -- (2.934,0.35);
    \node[lbl, anchor=west, orange!75!black] at (2.954,0.44999999999999996) {pace};
  \end{scope}
  \begin{scope}[shift={(5.55,0)}]
    \node[wpf] at (1.073,1.313) {}; \node[wpf] at (1.442,1.432) {};
    \draw[drf] plot[smooth] coordinates {(1.645,1.684) (1.680,1.697) (1.715,1.710) (1.749,1.724) (1.783,1.737) (1.816,1.752) (1.849,1.766) (1.881,1.781) (1.912,1.796) (1.944,1.812) (1.974,1.828) (2.004,1.844) (2.034,1.860) (2.063,1.877) (2.092,1.894) (2.120,1.912) (2.147,1.929) (2.174,1.947) (2.201,1.966) (2.227,1.984) (2.253,2.003) (2.278,2.022) (2.302,2.042) (2.326,2.061) (2.350,2.081) (2.373,2.102) (2.395,2.122) (2.418,2.143) (2.439,2.164) (2.460,2.185) (2.481,2.206) (2.501,2.228) (2.520,2.250) (2.539,2.272) (2.558,2.295) (2.576,2.317) (2.593,2.340) (2.610,2.363) (2.627,2.387) (2.643,2.411) (2.658,2.434) (2.673,2.459) (2.688,2.483) (2.702,2.507) (2.715,2.532) (2.728,2.557) (2.741,2.583) (2.753,2.608)};
    \node[box] at (1.977,1.804) {}; \node[box] at (2.165,2.035) {}; \node[box] at (2.431,2.180) {}; \node[box] at (2.606,2.390) {}; \node[box] at (2.814,2.554) {};
    \node[wp] at (1.645,1.684) {}; \node[wp] at (1.977,1.804) {}; \node[wp] at (2.165,2.035) {}; \node[wp] at (2.431,2.180) {}; \node[wp] at (2.606,2.390) {}; \node[wp] at (2.814,2.554) {};
    \fill[black!10] (0.350,0.18) rectangle (0.923,0.5800000000000001);
    \node[lbl, anchor=south, black!60] at (0.637,0.5800000000000001) {limit};
    \draw[->, black!70, line width=0.5pt] (0.24999999999999997,0.38) -- (3.75,0.38);
    \node[lbl, anchor=north east, black!70] at (3.75,0.34) {time};
    \draw[drf] (0.35,0.38) -- (2.508,0.38);
    \node[wps] at (0.350,0.38) {}; \node[wps] at (0.637,0.38) {}; \node[wps] at (0.923,0.38) {}; \node[wps] at (1.254,0.38) {}; \node[wps] at (1.792,0.38) {}; \node[wps] at (2.508,0.38) {};
  \end{scope}
  \begin{scope}[shift={(10.35,0)}]
    \draw[run] plot[smooth] coordinates {(0.390,0.985) (0.473,1.012) (0.556,1.038) (0.638,1.063) (0.721,1.087) (0.803,1.110) (0.886,1.133) (0.968,1.154) (1.051,1.174) (1.133,1.193) (1.216,1.212) (1.299,1.229) (1.381,1.245) (1.464,1.261) (1.546,1.275) (1.629,1.289)};
    \draw[runf, dashed] plot[smooth] coordinates {(1.629,1.289) (1.734,1.304) (1.840,1.318) (1.945,1.331) (2.050,1.342) (2.156,1.351) (2.261,1.359) (2.367,1.365) (2.472,1.369) (2.577,1.372) (2.683,1.374) (2.788,1.374) (2.894,1.372) (2.999,1.368) (3.105,1.363) (3.210,1.357) (3.315,1.348) (3.421,1.339)};
    \draw[drff] plot[smooth] coordinates {(1.645,1.684) (1.680,1.697) (1.715,1.710) (1.749,1.724) (1.783,1.737) (1.816,1.752) (1.849,1.766) (1.881,1.781) (1.912,1.796) (1.944,1.812) (1.974,1.828) (2.004,1.844) (2.034,1.860) (2.063,1.877) (2.092,1.894) (2.120,1.912) (2.147,1.929) (2.174,1.947) (2.201,1.966) (2.227,1.984) (2.253,2.003) (2.278,2.022) (2.302,2.042) (2.326,2.061) (2.350,2.081) (2.373,2.102) (2.395,2.122) (2.418,2.143) (2.439,2.164) (2.460,2.185) (2.481,2.206) (2.501,2.228) (2.520,2.250) (2.539,2.272) (2.558,2.295) (2.576,2.317) (2.593,2.340) (2.610,2.363) (2.627,2.387) (2.643,2.411) (2.658,2.434) (2.673,2.459) (2.688,2.483) (2.702,2.507) (2.715,2.532) (2.728,2.557) (2.741,2.583) (2.753,2.608)};
    \node[wpf] at (1.073,1.313) {}; \node[wpf] at (1.442,1.432) {};
    \draw[blue!70!black, line width=1pt] (1.670,1.629) -- (1.621,1.739); \draw[blue!70!black, line width=1pt] (1.989,1.767) -- (1.923,1.870); \draw[blue!70!black, line width=1pt] (2.263,1.934) -- (2.180,2.027); \draw[blue!70!black, line width=1pt] (2.493,2.128) -- (2.394,2.208); \draw[blue!70!black, line width=1pt] (2.676,2.346) -- (2.564,2.409); \draw[blue!70!black, line width=1pt] (2.814,2.586) -- (2.691,2.631);
    \draw[red!60, line width=1pt, densely dashed] plot[smooth] coordinates {(1.629,1.289) (1.664,1.295) (1.696,1.307) (1.722,1.326) (1.743,1.355) (1.758,1.394) (1.769,1.441) (1.777,1.493) (1.785,1.548) (1.794,1.603) (1.807,1.655) (1.825,1.701) (1.850,1.740) (1.880,1.772) (1.917,1.797) (1.956,1.818)};
    \node[lbl, anchor=south east, red!65] at ([xshift=-2pt,yshift=1pt]1.956,1.818) {late};
    \draw[tra] plot[smooth] coordinates {(1.629,1.289) (1.664,1.295) (1.698,1.304) (1.731,1.319) (1.764,1.341) (1.795,1.370) (1.826,1.408) (1.856,1.452) (1.885,1.501) (1.914,1.555) (1.943,1.611) (1.972,1.668) (2.001,1.723) (2.031,1.775) (2.062,1.823) (2.093,1.865) (2.125,1.901) (2.158,1.932) (2.190,1.958) (2.222,1.981)};
    \fill (1.629,1.289) circle (2pt);
    \node[lbl, anchor=north] at ([yshift=-3pt]1.629,1.289) {switch};
    \node[lbl, anchor=west, red!80!black] at ([xshift=4pt,yshift=-2pt]2.222,1.981) {on time};
    \draw[->, black!70, line width=0.5pt] (0.24999999999999997,0.38) -- (3.75,0.38);
    \node[lbl, anchor=north east, black!70] at (3.75,0.34) {time};
    \fill[black!45] (0.35,0.32) rectangle (1.274,0.44);
    \draw[tra] (1.274,0.38) -- (1.998,0.38);
    \draw[drf] (1.998,0.38) -- (3.583,0.38);
    \draw[blue!70!black, line width=1pt] (1.425,0.27) -- (1.425,0.49); \draw[blue!70!black, line width=1pt] (1.712,0.27) -- (1.712,0.49); \draw[blue!70!black, line width=1pt] (1.998,0.27) -- (1.998,0.49); \draw[blue!70!black, line width=1pt] (2.329,0.27) -- (2.329,0.49); \draw[blue!70!black, line width=1pt] (2.867,0.27) -- (2.867,0.49); \draw[blue!70!black, line width=1pt] (3.583,0.27) -- (3.583,0.49);
    \draw[red!65, line width=1pt, densely dashed] (1.274,0.6799999999999999) -- (1.862,0.6799999999999999);
    \draw[red!65, line width=0.8pt] (1.862,0.6099999999999999) -- (1.862,0.75);
    \draw[blue!70!black, densely dotted, line width=0.5pt] (1.712,0.49) -- (1.712,0.7799999999999999);
    \fill (1.274,0.38) circle (2pt);
  \end{scope}
  \begin{scope}[shift={(15.15,0)}]
    \draw[run] plot[smooth] coordinates {(0.390,0.985) (0.473,1.012) (0.556,1.038) (0.638,1.063) (0.721,1.087) (0.803,1.110) (0.886,1.133) (0.968,1.154) (1.051,1.174) (1.133,1.193) (1.216,1.212) (1.299,1.229) (1.381,1.245) (1.464,1.261) (1.546,1.275) (1.629,1.289)};
    \draw[runf, dashed] plot[smooth] coordinates {(1.629,1.289) (1.734,1.304) (1.840,1.318) (1.945,1.331) (2.050,1.342) (2.156,1.351) (2.261,1.359) (2.367,1.365) (2.472,1.369) (2.577,1.372) (2.683,1.374) (2.788,1.374) (2.894,1.372) (2.999,1.368) (3.105,1.363) (3.210,1.357) (3.315,1.348) (3.421,1.339)};
    \draw[white, line width=2.8pt] plot[smooth] coordinates {(1.629,1.289) (1.668,1.295) (1.707,1.302) (1.746,1.314) (1.784,1.331) (1.821,1.356) (1.858,1.388) (1.893,1.428) (1.928,1.475) (1.961,1.528) (1.994,1.586) (2.025,1.648) (2.056,1.711) (2.086,1.775) (2.116,1.837) (2.144,1.897) (2.173,1.953) (2.201,2.003) (2.228,2.048) (2.255,2.087) (2.282,2.120) (2.308,2.148) (2.333,2.171) (2.359,2.189) (2.383,2.205) (2.407,2.219) (2.430,2.231) (2.453,2.242) (2.475,2.252) (2.496,2.263) (2.517,2.273) (2.537,2.284) (2.556,2.296) (2.574,2.307) (2.592,2.319) (2.609,2.332) (2.626,2.345) (2.641,2.358) (2.657,2.372) (2.671,2.385) (2.685,2.399) (2.698,2.413) (2.711,2.426) (2.723,2.440) (2.735,2.454) (2.746,2.468) (2.756,2.481) (2.766,2.495) (2.776,2.509) (2.785,2.523) (2.793,2.537) (2.801,2.551) (2.808,2.564) (2.815,2.578) (2.818,2.585)};
    \draw[new, line width=1.5pt] plot[smooth] coordinates {(1.629,1.289) (1.668,1.295) (1.707,1.302) (1.746,1.314) (1.784,1.331) (1.821,1.356) (1.858,1.388) (1.893,1.428) (1.928,1.475) (1.961,1.528) (1.994,1.586) (2.025,1.648) (2.056,1.711) (2.086,1.775) (2.116,1.837) (2.144,1.897) (2.173,1.953) (2.201,2.003) (2.228,2.048) (2.255,2.087) (2.282,2.120) (2.308,2.148) (2.333,2.171) (2.359,2.189) (2.383,2.205) (2.407,2.219) (2.430,2.231) (2.453,2.242) (2.475,2.252) (2.496,2.263) (2.517,2.273) (2.537,2.284) (2.556,2.296) (2.574,2.307) (2.592,2.319) (2.609,2.332) (2.626,2.345) (2.641,2.358) (2.657,2.372) (2.671,2.385) (2.685,2.399) (2.698,2.413) (2.711,2.426) (2.723,2.440) (2.735,2.454) (2.746,2.468) (2.756,2.481) (2.766,2.495) (2.776,2.509) (2.785,2.523) (2.793,2.537) (2.801,2.551) (2.808,2.564) (2.815,2.578) (2.818,2.585)};
    \node[boxf] at (2.165,2.035) {}; \node[boxf] at (2.431,2.180) {}; \node[boxf] at (2.606,2.390) {}; \node[boxf] at (2.814,2.554) {};
    \fill (1.629,1.289) circle (2pt);
    \node[lbl, anchor=north] at ([yshift=-3pt]1.629,1.289) {switch};
    \node[lbl, anchor=north west] at ([xshift=2pt,yshift=-3pt]2.431,2.180) {new plan};
    \draw[->, black!70, line width=0.5pt] (0.24999999999999997,0.38) -- (3.75,0.38);
    \node[lbl, anchor=north east, black!70] at (3.75,0.34) {time};
    \fill[black!45] (0.35,0.32) rectangle (1.274,0.44);
    \fill[black] (1.274,0.32) rectangle (3.583,0.44);
    \node[draw=blue!60!black, fill=white, line width=0.6pt, minimum size=4.2pt, inner sep=0pt] at (1.998,0.38) {}; \node[draw=blue!60!black, fill=white, line width=0.6pt, minimum size=4.2pt, inner sep=0pt] at (2.329,0.38) {}; \node[draw=blue!60!black, fill=white, line width=0.6pt, minimum size=4.2pt, inner sep=0pt] at (2.867,0.38) {}; \node[draw=blue!60!black, fill=white, line width=0.6pt, minimum size=4.2pt, inner sep=0pt] at (3.583,0.38) {};
    \fill (1.274,0.38) circle (2pt);
  \end{scope}
\end{tikzpicture}}
\caption{One execution cycle. (1)~From an observation on the running plan, the policy predicts a new chunk and its pace. (2)~\textsc{Draft} fits a smooth path within boxes around the kept waypoints and times it at the pace unless a limit binds. (3)~\textsc{Splice} joins the running plan to this path at the earliest point it reaches on time. (4)~\textsc{Smooth} solves one minimum-jerk program through the boxes at their scheduled times, within all limits, giving the new plan.}
\label{fig:controller}
\end{figure}
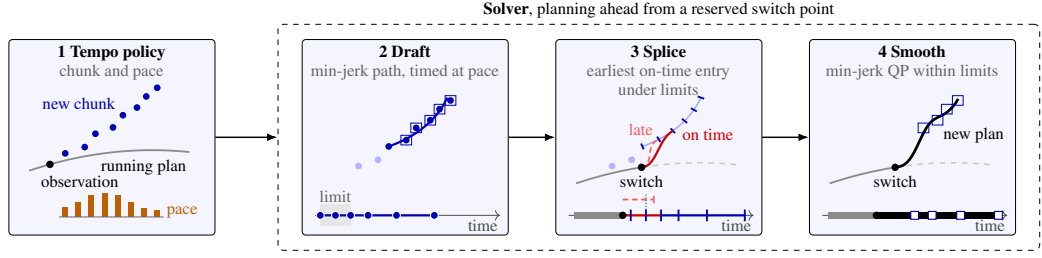

\subsection{Solver}
\label{sec:solver}

The solver runs three steps (Figure~\ref{fig:controller} and Algorithm~\ref{alg:controller}, with details in Appendix~\ref{app:controller}). \textsc{Draft} turns the new chunk into a timed path, \textsc{Splice} attaches this path to $P$ at the switch point, and \textsc{Smooth} turns the result into one plan within all limits.

\paragraph{Draft.}
\textsc{Draft} builds a timed path $\bm{\xi}$ from the kept waypoints of the new chunk alone. It first fits a minimum-jerk path $\bm{\gamma}(\sigma)$ through boxes of radius $\eta$ around them~\citep{flash1985coordination}. The path passes within each box rather than through each waypoint, which smooths out the jitter between predicted waypoints. It is parameterized by the base clock $\sigma$, on which step $h$ lies at $\sigma = h\delta$, so that its shape does not depend on the pace. It then plays the path back as fast as the velocity, acceleration and torque limits allow, subject to $\dot{\sigma} \le r_h$ on step $h$, by time-optimal path parameterization~\citep{pham2018new}. Each step thus runs at its target pace unless a limit binds, and then only that stretch slows down.

\paragraph{Splice.}
At $t_b$, $P$ is at $\mathbf{x}_b$ and moving, not at the start $\mathbf{w}_{h_0}$ of $\bm{\xi}$, so \textsc{Splice} connects the two with a short transition. It makes two choices. \emph{When}: \textsc{Splice} starts $\bm{\xi}$ at $t^\star$, and $\bm{\xi}(t)$ denotes its point due at time $t$ (the ticks in Figure~\ref{fig:controller}). The first kept waypoint $\mathbf{w}_{h_0}$ is thus due when $P$ reaches step $h_0$, and the rest of $\bm{\xi}$ follows on its own timing. \emph{Where}: \textsc{Splice} joins $\bm{\xi}$ at the earliest point that the arm reaches on time within its limits. Reaching $\mathbf{w}_{h_0}$ by $t^\star$ may be impossible, or may force the arm to brake hard or turn back, so later points of $\bm{\xi}$ are candidates as well. For each candidate entry $\bm{\xi}(t_e)$, Ruckig computes the fastest jerk-limited transition $\bm{\psi}_e$ from $\mathbf{s}_b$ that arrives there with the position, velocity and acceleration of $\bm{\xi}$~\citep{berscheid2021jerk}. With $D_e$ its duration, the transition arrives at $t_b + D_e$, and its lateness is $\ell_e = \max\big(0,\ t_b + D_e - t_e\big)$. \textsc{Splice} picks the entry with the smallest lateness and, among the entries reached on time ($\ell_e = 0$), the earliest whose transition passes within $\epsilon$ of every waypoint it skips, or the one that passes nearest if none does (Appendix~\ref{app:controller}). The arm thus joins $\mathbf{w}_{h_0}$ whenever it can do so on time, and otherwise cuts the corner near it and merges further along $\bm{\xi}$. The transition followed by the rest of $\bm{\xi}$ is the spliced reference $\bar{\mathbf{x}}$, on which each waypoint $\mathbf{w}_h$ after the entry is due at a time $t_h$.

\paragraph{Smooth.}
Neither piece respects all limits, as the transition bounds the torque only roughly and $\bm{\xi}$ does not bound the torque rate. \textsc{Smooth} thus solves a minimum-jerk optimal control problem on the timeline of $\bar{\mathbf{x}}$, as a quadratic program~\citep{frison2020hpipm}. It starts from $\mathbf{s}_b$ at $t_b$, keeps $\mathbf{x}(t_h)$ within $\eta$ of $\mathbf{w}_h$ for each waypoint after the entry, and enforces all limits, including the torque rate. The spliced reference only sets these times and the point about which the dynamics are linearized, and the program places no target on the transition itself. The resulting plan is checked against the limits before installation. If the program is infeasible or its plan fails the check, \textsc{Smooth} solves it again with the boxes relaxed to a penalty, and a plan that still fails is replaced by a braking plan.

\section{Experiments}
\label{sec:exp}

\begin{figure}[t]
\centering
\includegraphics[width=\linewidth]{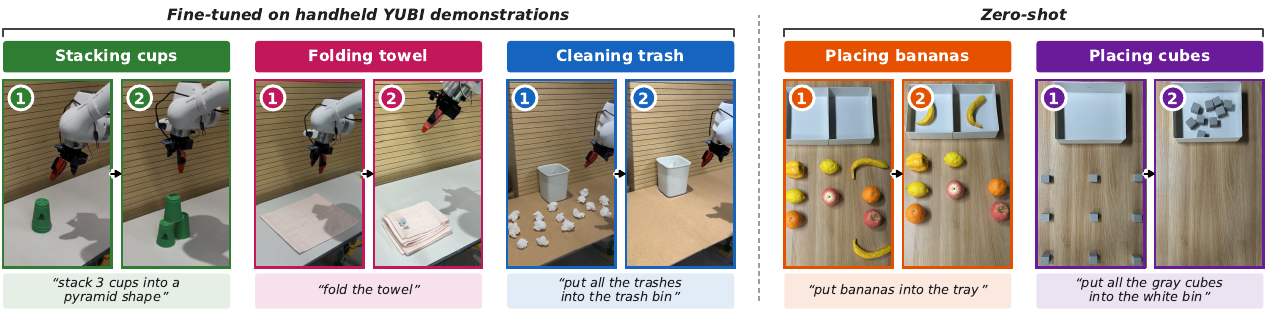}
\caption{Real-world tasks and their instructions, each shown at its initial scene (1) and its goal (2). The policies for stacking cups, folding towel and cleaning trash are fine-tuned on handheld YUBI demonstrations, and placing bananas and placing cubes are performed zero-shot (Appendix~\ref{app:zero-shot-eval}).}
\label{fig:real_world_tasks}
\end{figure}

\subsection{Training pipeline}
\label{sec:pipeline}

Our experiments have two stages. In both, our policy learns the median clock and the tempo channels of the 0.9, 0.95 and 0.99 quantiles (Section~\ref{sec:policy}), and every policy uses train-time RTC~\citep{black2025training}. In the first stage, we train the speed prior on the handheld demonstrations of FastUMI-100K and the teleoperated ones of DROID, retime all DROID episodes with the FastUMI-100K label, and fine-tune the released $\pi_{0.5}$ DROID policy on them~\citep{khazatsky2024droid,liu2025fastumi,intelligence2025openpi}. We evaluate this policy in simulation on RoboLab (Section~\ref{sec:robolab}) and zero-shot on the real robot (Section~\ref{sec:real_robot}). In the second stage, we adapt both models to handheld demonstrations collected with a YUBI gripper~\citep{ohkawa2026yubi}. For each task, we fine-tune the speed prior with the shortest quarter of episodes as the target domain and retime all demonstrations with the target label. We then fine-tune the first-stage policy on the retimed demonstrations and evaluate it on the real robot (Section~\ref{sec:real_robot}).

\subsection{Simulation on RoboLab}
\label{sec:robolab}

\paragraph{Tasks and metrics.}
RoboLab~\citep{yang2026robolab} is a high-fidelity benchmark for task-generalist policies in simulation. Its 120 new language-conditioned tasks span stacking, pick-and-place, tool use and rearrangement, and we run 10 episodes per task. An episode fails if it exceeds the time limit of its task, so running faster alone can raise the success rate. We therefore compare policies at matched speed-up, the mean end-effector speed relative to the default policy at $1\times$. We report the success rate (SR) and the throughput with regret (TPR) of \citet{arachchige2025sail}, which favors policies that are both fast and reliable (Appendix~\ref{app:real_robot}).

\paragraph{Baselines.}
We compare our policy with two policies trained on the recorded timing of DROID. The default policy is trained on all episodes. Filtered behavior cloning (Filtered BC) is trained only on the fastest 16,609 of 74,546 episodes, whose pace of $1.5\times$ matches our median clock, and tests whether imitating fast demonstrations suffices. Methods that find the segments to accelerate need a proxy policy or labeled demonstrations of the task at hand~\citep{guo2025demospeedup,kim2026espada}, whereas our policy is trained on many tasks and evaluated on unseen ones. We therefore compare with uniform acceleration, which applies to any policy: the default policy plays at $1.5\times$, $2\times$ or $2.5\times$, and our policy on its base clock and on each tempo channel. In simulation, all policies play their chunks by interpolating the joint targets, without the controller of Section~\ref{sec:controller}, so the comparison isolates the clocks (Appendix~\ref{app:robolab}).

\begin{table}[t]
\centering
\caption{Throughput and motion quality at matched speed-up on RoboLab. TPR is in successes per minute, Peak is the peak end-effector speed in \unit{m/s} and Jerk the integrated squared jerk in $10^3\,\unit{m^2/s^5}$. SPARC, Peak and Jerk use the episodes that every execution of the baseline and our policy in Figure~\ref{fig:speed_success} completes.}
\label{tab:robolab_motion}
\begin{tabular}{@{}lcccccc@{}}
  \toprule
  Execution & Speed-up & SR$\uparrow$ & TPR$\uparrow$ & SPARC$\uparrow$ & Peak$\downarrow$ & Jerk$\downarrow$ \\
  \midrule
  Uniform $1\times$     & 1.00 & 0.462 & 1.45 & \textminus3.62 & 0.30 & 0.31 \\
  \midrule
  Uniform $2\times$     & 1.90 & 0.533 & 3.55 & \textminus3.87 & 0.52 & 1.23 \\
  Ours, $q = 0.95$      & 2.00 & \textbf{0.577} & \textbf{4.39} & \textbf{\textminus3.14} & \textbf{0.44} & \textbf{0.83} \\
  \midrule
  Uniform $2.5\times$   & 2.48 & 0.530 & 3.86 & \textminus4.11 & 0.69 & 2.41 \\
  Ours, $q = 0.99$      & 2.38 & \textbf{0.608} & \textbf{5.16} & \textbf{\textminus3.26} & \textbf{0.52} & \textbf{1.16} \\
  \bottomrule
\end{tabular}
\end{table}

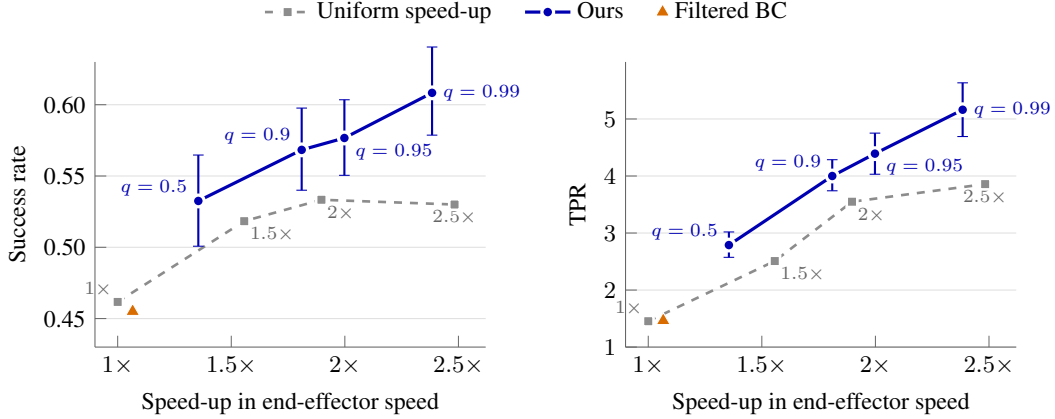
\begin{figure}[t]
\centering
\pgfplotslegendfromname{leg:speed_success}\\[1pt]
\begin{tikzpicture}
\begin{axis}[
  scale only axis, width=0.37\linewidth, height=0.27\linewidth,
  xmin=0.9, xmax=2.62, ymin=43, ymax=63,
  xtick={1,1.5,2,2.5}, xticklabels={$1\times$,$1.5\times$,$2\times$,$2.5\times$},
  ytick={45,50,55,60}, yticklabels={0.45,0.50,0.55,0.60},
  xlabel={Speed-up in end-effector speed}, ylabel={Success rate},
  ymajorgrids, grid style={black!12},
  axis lines*=left, axis line style={black!60}, tick style={black!60},
  tick label style={font=\footnotesize}, label style={font=\small},
  legend to name=leg:speed_success, legend columns=3, legend style={draw=none, font=\footnotesize, /tikz/every even column/.append style={column sep=10pt}},
  legend cell align=left, clip=false,
]
\addplot[black!45, dashed, line width=1pt, mark=square*, mark size=2pt, mark options={solid, fill=black!45, draw=white}]
  coordinates {(1.000,46.167) (1.557,51.833) (1.897,53.333) (2.483,53.000)};
\addlegendentry{Uniform speed-up}
\addplot[blue!70!black, line width=1.2pt, mark=*, mark size=2.2pt, mark options={fill=blue!70!black, draw=white},
  error bars/.cd, y dir=both, y explicit, error bar style={line width=0.8pt, blue!70!black}, error mark options={rotate=90, mark size=2pt, blue!70!black}]
  coordinates {(1.355,53.250) += (0,3.221) -= (0,3.175) (1.810,56.833) += (0,2.932) -= (0,2.830) (1.998,57.667) += (0,2.682) -= (0,2.619) (2.384,60.833) += (0,3.216) -= (0,2.966)};
\addlegendentry{Ours}
\addplot[only marks, orange!85!black, mark=triangle*, mark size=3pt, mark options={fill=orange!85!black, draw=white}]
  coordinates {(1.066,45.500)};
\addlegendentry{Filtered BC}
\node[font=\scriptsize, black!60, anchor=south east, inner sep=2pt] at (axis cs:1.000,46.167) {$1\times$};
\node[font=\scriptsize, black!60, anchor=north west, inner sep=2pt] at (axis cs:1.557,51.833) {$1.5\times$};
\node[font=\scriptsize, black!60, anchor=north west, inner sep=2pt] at (axis cs:1.897,53.333) {$2\times$};
\node[font=\scriptsize, black!60, anchor=north, inner sep=2pt] at (axis cs:2.483,53.000) {$2.5\times$};
\node[font=\scriptsize, blue!70!black, anchor=south east, xshift=-2pt, inner sep=2pt] at (axis cs:1.355,53.250) {$q = 0.5$};
\node[font=\scriptsize, blue!70!black, anchor=south east, xshift=-2pt, inner sep=2pt] at (axis cs:1.810,56.833) {$q = 0.9$};
\node[font=\scriptsize, blue!70!black, anchor=north west, xshift=2pt, inner sep=2pt] at (axis cs:1.998,57.667) {$q = 0.95$};
\node[font=\scriptsize, blue!70!black, anchor=west, xshift=2pt, inner sep=2pt] at (axis cs:2.384,60.833) {$q = 0.99$};
\end{axis}
\end{tikzpicture}
\hfill
\begin{tikzpicture}
\begin{axis}[
  scale only axis, width=0.37\linewidth, height=0.27\linewidth,
  xmin=0.9, xmax=2.62, ymin=1, ymax=6,
  xtick={1,1.5,2,2.5}, xticklabels={$1\times$,$1.5\times$,$2\times$,$2.5\times$},
  ytick={1,2,3,4,5},
  xlabel={Speed-up in end-effector speed}, ylabel={TPR},
  ymajorgrids, grid style={black!12},
  axis lines*=left, axis line style={black!60}, tick style={black!60},
  tick label style={font=\footnotesize}, label style={font=\small},
  legend style={draw=none, fill=none, font=\footnotesize, at={(0.98,0.04)}, anchor=south east, cells={anchor=west}},
  legend cell align=left, clip=false,
]
\addplot[black!45, dashed, line width=1pt, mark=square*, mark size=2pt, mark options={solid, fill=black!45, draw=white}]
  coordinates {(1.000,1.454) (1.557,2.509) (1.897,3.549) (2.483,3.858)};
\addplot[blue!70!black, line width=1.2pt, mark=*, mark size=2.2pt, mark options={fill=blue!70!black, draw=white},
  error bars/.cd, y dir=both, y explicit, error bar style={line width=0.8pt, blue!70!black}, error mark options={rotate=90, mark size=2pt, blue!70!black}]
  coordinates {(1.355,2.789) += (0,0.230) -= (0,0.215) (1.810,4.000) += (0,0.286) -= (0,0.259) (1.998,4.392) += (0,0.360) -= (0,0.361) (2.384,5.162) += (0,0.472) -= (0,0.470)};
\addplot[only marks, orange!85!black, mark=triangle*, mark size=3pt, mark options={fill=orange!85!black, draw=white}]
  coordinates {(1.066,1.465)};
\node[font=\scriptsize, black!60, anchor=south east, inner sep=2pt] at (axis cs:1.000,1.454) {$1\times$};
\node[font=\scriptsize, black!60, anchor=north west, inner sep=2pt] at (axis cs:1.557,2.509) {$1.5\times$};
\node[font=\scriptsize, black!60, anchor=north west, inner sep=2pt] at (axis cs:1.897,3.549) {$2\times$};
\node[font=\scriptsize, black!60, anchor=north, inner sep=2pt] at (axis cs:2.483,3.858) {$2.5\times$};
\node[font=\scriptsize, blue!70!black, anchor=south east, xshift=-2pt, inner sep=2pt] at (axis cs:1.355,2.789) {$q = 0.5$};
\node[font=\scriptsize, blue!70!black, anchor=south east, xshift=-2pt, inner sep=2pt] at (axis cs:1.810,4.000) {$q = 0.9$};
\node[font=\scriptsize, blue!70!black, anchor=north west, xshift=2pt, inner sep=2pt] at (axis cs:1.998,4.392) {$q = 0.95$};
\node[font=\scriptsize, blue!70!black, anchor=west, xshift=2pt, inner sep=2pt] at (axis cs:2.384,5.162) {$q = 0.99$};
\end{axis}
\end{tikzpicture}
\caption{Success rate (left) and TPR (right) against speed-up on RoboLab. Gray labels give the commanded speed-up. TPR (Appendix~\ref{app:real_robot}) is in successes per minute, with the time limit of each task. Bars are 95\% intervals of the gain over the uniform curve, bootstrapped over tasks.}
\label{fig:speed_success}
\end{figure}

\paragraph{Results.}
Uniform acceleration raises the SR of the default policy from 46.2\% at $1\times$ to 53.3\% at $1.9\times$ but no further (Figure~\ref{fig:speed_success}). Our policy lies above this curve on all four clocks, on average by 4.9 percentage points in SR and 0.88 successes per minute in TPR (95\% CIs [2.8, 7.1] and [0.63, 1.15]). Filtered BC runs only $1.07\times$ faster than the default policy and succeeds no more often, although its demonstrations match our median clock. At $2.4\times$, our policy moves more smoothly than uniform acceleration to $2.5\times$ (Table~\ref{tab:robolab_motion}): the spectral arc length (SPARC) of its end-effector speed is $-3.26$ against $-4.11$, and its peak speed and jerk are 24\% and 52\% lower~\citep{balasubramanian2015analysis}.

\subsection{Real-robot experiments}
\label{sec:real_robot}

\paragraph{Tasks and metrics.}
With one YUBI gripper~\citep{ohkawa2026yubi}, we collect demonstrations of three tasks by hand: 412 episodes of stacking cups, 242 of folding a towel and 61 of cleaning trash (Figure~\ref{fig:real_world_tasks}). The tracked gripper poses are mapped by inverse kinematics to joint positions of the Franka Research 3, which carries the same gripper. The shortest quarter of the episodes of each task forms the target domain, with 98, 58 and 15 episodes and 16\%, 16\% and 20\% of the frames. A cup or towel episode succeeds if the three cups form a pyramid or the towel is folded within \qty{10}{s}. On these two tasks, we run 30 episodes per condition and report SR, the average time of the successful rollouts (ATR), TPR and SPARC (Appendix~\ref{app:real_robot}). In cleaning trash, the robot has \qty{60}{s} to put 15 items into the bin, and we report the items placed per hour (UPH) and SPARC over 5 episodes.

\paragraph{Baselines.}
We compare our policy with the default policy, trained on all episodes, and Filtered BC, trained only on the target episodes, both on their recorded timing and fine-tuned from the same first-stage checkpoint as our policy. Unlike in simulation, all policies run under the controller of Section~\ref{sec:controller}: the baselines at $\alpha = 1$, and our policy at $\alpha = 1$ and on each of its tempo channels q90, q95 and q99. To measure the effect of the controller, the default policy also runs under the PD baseline of Appendix~\ref{app:controller}, which holds each action as a position target for one step at \qty{15}{Hz}, as the openpi DROID example does~\citep{intelligence2025openpi}.

\paragraph{Results.}
Table~\ref{tab:real_robot_yubi} reports the results. In stacking cups, the default policy takes \qty{6.02}{s}. Our policy takes \qty{4.40}{s} on its base clock and between \qty{3.25}{s} and \qty{3.65}{s} on its tempo channels, at success rates of 0.93 to 0.97. On the 0.95 tempo channel, it succeeds as often as the default policy, in 29 of 30 episodes, but takes only \qty{3.50}{s}, raising TPR from 9.45 to 16.38. For reference, SAIL reports \qty{14.71}{s} at a success rate of 0.40 on the same task, accelerating a diffusion policy trained on 50 teleoperated demonstrations to $1.76\times$ their pace, although on a different robot~\citep{arachchige2025sail}. Our policy is thus $4.2\times$ faster and succeeds more than twice as often. Filtered BC succeeds in only 9 of 30 episodes. In folding towel, the tempo channels shorten the time from \qty{5.52}{s} to between \qty{3.85}{s} and \qty{4.39}{s} at success rates of 0.97 to 1.00. In cleaning trash, they raise UPH from 444 to up to 809.

Under the PD baseline, the default policy succeeds in 28 and only 4 of 30 episodes on cups and towel and places no trash items, against 29, 29 and 444 items per hour under our controller. On towel, our policy also moves more smoothly than the default policy on every clock, with a SPARC between $-3.29$ and $-3.04$ against $-3.75$. On cups and trash, its SPARC stays close to that of the default policy, between $-3.00$ and $-2.70$ against $-2.68$ and between $-5.17$ and $-4.49$ against $-4.82$.

\paragraph{Zero-shot transfer.}
Without fine-tuning, we also run the default and our policies of Section~\ref{sec:robolab} on this robot with the Robotiq gripper of DROID, on placing bananas and placing cubes (Figure~\ref{fig:real_world_tasks}). On its 0.99 tempo channel, our policy places both bananas in all 30 episodes and all nine cubes in 12, whereas the default policy under our controller does so in 22 and none. UPH rises from 232 to 493 on bananas and from 354 to 511 on cubes (Appendix~\ref{app:zero-shot-eval}). Under the PD baseline, the default policy places both bananas in only 17 episodes and 3.90 cubes on average, and our policy moves most smoothly on its base clock in both tasks.

\begin{table}[hbt]
    \centering
    \setlength{\tabcolsep}{3.5pt}
    \caption{Real-robot results on the YUBI tasks of Figure~\ref{fig:real_world_tasks}. Ours (1x) runs our controller at $\alpha = 1$, and Ours (q90), (q95) and (q99) on the tempo channels of the 0.9, 0.95 and 0.99 quantiles. TPR is in successes per minute, ATR in seconds and UPH in items per hour. Appendix~\ref{app:real_robot} defines the metrics.}
    \label{tab:real_robot_yubi}
    \resizebox{\columnwidth}{!}{
    \begin{tabular}{llcccccccccc}
    \toprule
             &             &   \multicolumn{4}{c}{Stacking cups} & \multicolumn{4}{c}{Folding towel} & \multicolumn{2}{c}{Cleaning trash} \\
             \cmidrule(l){3-6} \cmidrule(l){7-10} \cmidrule(l){11-12}
       Model & Controller  & SR$\uparrow$ & TPR$\uparrow$ & ATR$\downarrow$ & SPARC$\uparrow$ & SR$\uparrow$ & TPR$\uparrow$ & ATR$\downarrow$ & SPARC$\uparrow$ & UPH$\uparrow$ & SPARC$\uparrow$ \\
       \midrule
       Default & PD            & 0.93 & 8.25 & 6.48 & \textminus2.79 & 0.13 & \textminus4.02 & 6.81 & \textminus3.46 & 0 & \textminus5.41 \\
       Default & Ours (1x)     & \textbf{0.97} & 9.45 & 6.02 & \textbf{\textminus2.68} & 0.97 & 10.31 & 5.52 & \textminus3.75 & 444 & \textminus4.82 \\
       Filtered BC & Ours (1x) & 0.30 & \textminus1.59 & 7.14 & \textminus3.80 & 0.60 & 4.25 & 5.41 & \textminus3.46 & 348 & \textbf{\textminus4.35} \\
       Ours & Ours (1x)        & \textbf{0.97} & 12.99 & 4.40 & \textminus2.70 & 0.97 & 10.59 & 5.38 & \textminus3.29 & 600 & \textminus4.49 \\
       Ours & Ours (q90)       & \textbf{0.97} & 15.68 & 3.65 & \textminus2.72 & 0.97 & 13.01 & 4.39 & \textminus3.19 & 732 & \textminus5.17 \\
       Ours & Ours (q95)       & \textbf{0.97} & 16.38 & 3.50 & \textminus2.82 & 0.97 & 13.62 & 4.20 & \textminus3.14 & 792 & \textminus4.85 \\
       Ours & Ours (q99)       & 0.93 & \textbf{16.84} & \textbf{3.25} & \textminus3.00 & \textbf{1.00} & \textbf{15.61} & \textbf{3.85} & \textbf{\textminus3.04} & \textbf{809} & \textminus4.84 \\
    \bottomrule
    \end{tabular}
    }
\end{table}

\subsection{Analysis of the speed prior}
\label{sec:prior_analysis}

We analyze the speed prior of the first stage on held-out validation episodes, reading each domain with its own label unless stated otherwise. Appendix~\ref{app:prior_analysis} gives the data and further results.

\paragraph{Pace within an episode.}
To slow down where demonstrators do, such as at a grasp, the prior must predict how the pace changes within a demonstration, not only how fast the demonstration is overall. We isolate these changes by subtracting each episode's mean log speed from both the recorded and the predicted speeds. The predicted median explains 94\% of what remains on DROID and 92\% on FastUMI-100K (Table~\ref{tab:prior_calibration}).

\paragraph{Transfer to the handheld pace.}
We next examine which parts of a DROID path become faster when the prior reads it with the FastUMI-100K label. The arm speeds up more next to grasps and releases than in transit, by $1.57\times$ against $1.28\times$ at the median, because handheld demonstrators slow down less around gripper events than teleoperators do (Table~\ref{tab:prior_phase}). The transferred clock thus shortens the hesitation of teleoperation rather than accelerating the motion uniformly.

\paragraph{Smoothness of the retimed demonstrations.}
Finally, we examine where the smoothness of our policies in Table~\ref{tab:robolab_motion} comes from, by comparing held-out DROID episodes before and after retiming. Retiming with the median clock raises SPARC from $-3.91$ to $-3.09$, whereas uniform acceleration to the same duration leaves it at $-3.87$, since it keeps the stops and hesitations of the recording. The smoothness is therefore already present in the demonstrations that our policies are trained on.

\section{Conclusion}

We learned a speed prior that predicts how fast demonstrators traverse each part of an end-effector path, and used it to give demonstrations the pace of faster ones, those recorded with a handheld gripper or the shortest quarter of the episodes of a task. Trained once across tasks, the prior retimes demonstrations whose paths the faster demonstrators never followed. Training only on the faster demonstrations does not yield a policy that is both fast and reliable, whereas giving their pace to all demonstrations does. At the same speed-up, the resulting policies succeed more often than uniformly accelerated policies and move more smoothly, and our controller executes their pace within the limits of a real robot. One limitation is that the prior only transfers paces that some demonstrators have shown, so it cannot speed up motion beyond what is observed in the demonstrations.

\clearpage

\subsection*{AI use statement}

We used AI coding assistants (Claude Code and Codex) to help implement our
methods: the authors designed the method, including its mathematical
formulation and architecture, and the tools generated code according to our
specifications. We also used generative AI tools to polish the writing of
author-drafted text. We have not used generative AI tools for any other task
requiring disclosure (e.g., generating data, developing theory or proofs,
proposing hypotheses, designing experiments, translation, data analysis, or
interpreting results). All AI-generated code was reviewed line by line and
verified for correctness by the authors, and all AI-assisted edits to the text
were checked to ensure they did not change the technical content. We take
responsibility for the final content of this work, including text, claims or
artifacts produced with the aid of generative AI.

\subsection*{Ethics statement}

We use publicly available datasets (e.g., DROID and FastUMI) under their
respective licenses, along with handheld demonstration data collected by members
of our research team that contain no personally identifiable information. All
real-robot experiments were conducted in a controlled laboratory environment
with appropriate safety measures. As learned policies may behave unpredictably
in unseen settings, real-world deployment near humans would require additional
safety validation.



\bibliography{iclr2027_conference}

@inproceedings{guo2025demospeedup,
title={DemoSpeedup: Accelerating visuomotor policies via entropy-guided demonstration acceleration},
author={Guo, Lingxiao and Xue, Zhengrong and Xu, Zijing and Xu, Huazhe},
booktitle={CoRL},
year={2025}
}

@article{kim2026espada,
title={Espada: Execution speedup via semantics aware demonstration data downsampling for imitation learning},
author={Kim, Byung-ju and Pahk, Jinu and Lee, Chungwoo and Kim, Jaejoon and Lee, Jangha and Kim, Theo Taeyeong and Shim, Kyuhwan and Lee, Jun Ki and Zhang, Byoung-Tak},
journal={IEEE Robotics and Automation Letters},
year={2026}
}

@article{sanchez2026volt,
title={VOLT: Vision and Language Trajectory Segmentation for Faster-than-Demonstration Policies},
author={Sanchez, Robert Ramirez and Evans, Daniel J and Losey, Dylan P and Jain, Siddarth},
journal={arXiv preprint arXiv:2606.06323},
year={2026}
}

@inproceedings{hu2026autospeed,
title={AutoSpeed: Annotation-Free Stage-Adaptive Motion Speed Learning for Robot Manipulation},
author={Hu, Qingda and Qiu, Ziheng and Zhao, Jieru and Gan, Zhongxue and Ding, Wenchao},
booktitle={ECCV},
year={2026}
}

@article{jing2026tempovla,
title={Tempovla: Learning speed-controllable vision-language-action policies},
author={Jing, Dong and Nie, Jingchen and Zhang, Tianqi and Liu, Jiaqi and Yao, Huaxiu and Lu, Zhiwu and Ding, Mingyu},
journal={arXiv preprint arXiv:2606.06491},
year={2026}
}

@inproceedings{arachchige2025sail,
title={Sail: Faster-than-demonstration execution of imitation learning policies},
author={Arachchige, Nadun Ranawaka and Chen, Zhenyang and Jung, Wonsuhk and Shin, Woo Chul and Bansal, Rohan and Barroso, Pierre and He, Yu Hang and Lin, Yingyan Celine and Joffe, Benjamin and Kousik, Shreyas and others},
booktitle={CoRL},
year={2025}
}

@article{nam2025speedaug,
title={Speedaug: Policy acceleration via tempo-enriched policy and rl fine-tuning},
author={Nam, Taewook and Cho, Junmo and Jang, Youngsoo and Hwang, Sung Ju},
journal={arXiv preprint arXiv:2512.00062},
year={2025}
}

@inproceedings{kim2026time,
title={Time Optimal Execution of Action Chunk Policies Beyond Demonstration Speed},
author={Sunwoo Kim and Jeongjun Kim and Joseph J Lim},
booktitle={ICLR},
year={2026},
}

@inproceedings{braglia2025arc,
title={Arc-length-based warping for robot skill synthesis from multiple demonstrations},
author={Braglia, Giovanni and Tebaldi, Davide and Lazzaretti, Andr{\'e} E and Biagiotti, Luigi},
booktitle={IROS},
year={2025},
}

@article{yang2026improving,
title={Improving Robotic Imitation Learning via Trajectory Standardization},
author={Yang, Licheng and Qian, Lingfeng and Zheng, Fei and He, Yonghao and Sui, Wei and Li, Shuangshuang and Su, Hu},
journal={arXiv preprint arXiv:2606.22907},
year={2026}
}

@inproceedings{intelligence2025pi05,
title={$\pi_{0.5}$: a Vision-Language-Action Model with Open-World Generalization},
author={Intelligence, Physical and Black, Kevin and Brown, Noah and Darpinian, James and Dhabalia, Karan and Driess, Danny and Esmail, Adnan and Equi, Michael and Finn, Chelsea and Fusai, Niccolo and others},
booktitle={CoRL},
year={2025}
}

@article{pham2018new,
title={A new approach to time-optimal path parameterization based on reachability analysis},
author={Pham, Hung and Pham, Quang-Cuong},
journal={IEEE Transactions on Robotics},
year={2018}
}

@article{park1995distance,
title={Distance Metrics on the Rigid-Body Motions with Applications to Mechanism Design},
author={Frank Chongwoo Park},
journal={Journal of Mechanical Design},
year={1995}
}

@inproceedings{verduyn2024enhancing,
title={Enhancing motion trajectory segmentation of rigid bodies using a novel screw-based trajectory-shape representation},
author={Verduyn, Arno and Vochten, Maxim and De Schutter, Joris},
booktitle={ICRA},
year={2024}
}

@article{chi2024universal,
title={Universal manipulation interface: In-the-wild robot teaching without in-the-wild robots},
author={Chi, Cheng and Xu, Zhenjia and Pan, Chuer and Cousineau, Eric and Burchfiel, Benjamin and Feng, Siyuan and Tedrake, Russ and Song, Shuran},
journal={arXiv preprint arXiv:2402.10329},
year={2024}
}

@inproceedings{farebrother2024stop,
title={Stop regressing: Training value functions via classification for scalable deep rl},
author={Farebrother, Jesse and Orbay, Jordi and Vuong, Quan and Ta{\"\i}ga, Adrien Ali and Chebotar, Yevgen and Xiao, Ted and Irpan, Alex and Levine, Sergey and Castro, Pablo Samuel and Faust, Aleksandra and others},
booktitle={ICML},
year={2024}
}

@article{zhang2025qwen3,
title={Qwen3 embedding: Advancing text embedding and reranking through foundation models},
author={Zhang, Yanzhao and Li, Mingxin and Long, Dingkun and Zhang, Xin and Lin, Huan and Yang, Baosong and Xie, Pengjun and Yang, An and Liu, Dayiheng and Lin, Junyang and others},
journal={arXiv preprint arXiv:2506.05176},
year={2025}
}

@article{khazatsky2024droid,
title={Droid: A large-scale in-the-wild robot manipulation dataset},
author={Khazatsky, Alexander and Pertsch, Karl and Nair, Suraj and Balakrishna, Ashwin and Dasari, Sudeep and Karamcheti, Siddharth and Nasiriany, Soroush and Srirama, Mohan Kumar and Chen, Lawrence Yunliang and Ellis, Kirsty and others},
journal={arXiv preprint arXiv:2403.12945},
year={2024}
}

@article{zhao2025vla,
title={{VLA-RAIL}: A Real-Time Asynchronous Inference Linker for {VLA} Models and Robots},
author={Zhao, Yongsheng and Zhao, Lei and Cheng, Baoping and Yao, Gongxin and Wen, Xuanzhang and Gao, Han},
journal={arXiv preprint arXiv:2512.24673},
year={2025}
}

@inproceedings{berscheid2021jerk,
title={Jerk-limited Real-time Trajectory Generation with Arbitrary Target States},
author={Berscheid, Lars and Kr{\"o}ger, Torsten},
booktitle={RSS},
year={2021}
}

@article{frison2020hpipm,
title={{HPIPM}: a high-performance quadratic programming framework for model predictive control},
author={Frison, Gianluca and Diehl, Moritz},
journal={IFAC-PapersOnLine},
year={2020}
}

@inproceedings{black2025real,
title={Real-Time Execution of Action Chunking Flow Policies},
author={Black, Kevin and Galliker, Manuel Y and Levine, Sergey},
booktitle={NeurIPS},
year={2025}
}

@article{black2025training,
title={Training-Time Action Conditioning for Efficient Real-Time Chunking},
author={Black, Kevin and Ren, Allen Z and Equi, Michael and Levine, Sergey},
journal={arXiv preprint arXiv:2512.05964},
year={2025}
}

@article{flash1985coordination,
title={The coordination of arm movements: an experimentally confirmed mathematical model},
author={Flash, Tamar and Hogan, Neville},
journal={Journal of Neuroscience},
volume={5},
number={7},
pages={1688--1703},
year={1985}
}

@misc{lin2021polymetis,
title={Polymetis},
author={Lin, Yixin and Wang, Austin S. and Sutanto, Giovanni and Rai, Akshara and Meier, Franziska},
howpublished={\url{https://facebookresearch.github.io/fairo/polymetis/}},
year={2021}
}

@misc{intelligence2025openpi,
title={openpi},
author={{Physical Intelligence}},
howpublished={\url{https://github.com/Physical-Intelligence/openpi}},
year={2025}
}

@article{ohkawa2026yubi,
title={{YUBI}: Yielding Universal Bidigital Interface for Bimanual Dexterous Manipulation at Scale},
author={Ohkawa, Takehiko and Arima, Jumpei and Noguchi, Yuki and Tateno, Masatoshi and Sugiura, Makoto and Okubo, Takuya and Ikeuchi, Kengo and Shin, Yuma and Nishizawa, Hiroki and Kanazawa, Naoaki and others},
journal={arXiv preprint arXiv:2606.10244},
year={2026}
}

@article{yang2026robolab,
title={Robolab: A high-fidelity simulation benchmark for analysis of task generalist policies},
author={Yang, Xuning and Dagli, Rishit and Zook, Alex and Hadfield, Hugo and Goyal, Ankit and Birchfield, Stan and Ramos, Fabio and Tremblay, Jonathan},
journal={arXiv preprint arXiv:2604.09860},
year={2026}
}

@article{balasubramanian2015analysis,
title={On the analysis of movement smoothness},
author={Balasubramanian, Sivakumar and Melendez-Calderon, Alejandro and Roby-Brami, Agnes and Burdet, Etienne},
journal={Journal of neuroengineering and rehabilitation},
year={2015}
}

@article{liu2025fastumi,
title={Fastumi-100k: Advancing data-driven robotic manipulation with a large-scale umi-style dataset},
author={Liu, Kehui and Jia, Zhongjie and Li, Yang and Zhaxizhuoma and Chen, Pengan and Liu, Song and Liu, Xin and Zhang, Pingrui and Song, Haoming and Ye, Xinyi  and others},
journal={arXiv preprint arXiv:2510.08022},
year={2025}
}

@article{yang2026realtime,
title={{Realtime-VLA V2}: Learning to Run {VLAs} Fast, Smooth, and Accurate},
author={Yang, Chen and Hu, Yucheng and Ma, Yunchao and Yang, Yunhuan and Tan, Jing and Fan, Haoqiang},
journal={arXiv preprint arXiv:2603.26360},
year={2026}
}

@inproceedings{martinmartin2019variable,
title={Variable Impedance Control in End-Effector Space: An Action Space for Reinforcement Learning in Contact-Rich Tasks},
author={Mart{\'\i}n-Mart{\'\i}n, Roberto and Lee, Michelle A and Gardner, Rachel and Savarese, Silvio and Bohg, Jeannette and Garg, Animesh},
booktitle={IROS},
year={2019}
}

@inproceedings{kamijo2024learning,
title={Learning Variable Compliance Control From a Few Demonstrations for Bimanual Robot with Haptic Feedback Teleoperation System},
author={Kamijo, Tatsuya and Beltran-Hernandez, Cristian C and Hamaya, Masashi},
booktitle={IROS},
year={2024}
}

@inproceedings{hou2025adaptive,
title={Adaptive Compliance Policy: Learning Approximate Compliance for Diffusion Guided Control},
author={Hou, Yifan and Liu, Zeyi and Chi, Cheng and Cousineau, Eric and Kuppuswamy, Naveen and Feng, Siyuan and Burchfiel, Benjamin and Song, Shuran},
booktitle={ICRA},
year={2025}
}

@inproceedings{kim2026compvla,
title={{CompVLA}: A Variable Compliance Vision-Language-Action Model for Contact-rich Manipulation},
author={Kim, Jongmin and Ha, Junsu and Park, Che-Sang and others},
booktitle={CoRL},
year={2026}
}

@article{simeoni2025dinov3,
  title={Dinov3},
  author={Sim{\'e}oni, Oriane and Vo, Huy V and Seitzer, Maximilian and Baldassarre, Federico and Oquab, Maxime and Jose, Cijo and Khalidov, Vasil and Szafraniec, Marc and Yi, Seungeun and Ramamonjisoa, Micha{\"e}l and others},
  journal={arXiv preprint arXiv:2508.10104},
  year={2025}
}

@article{pertsch2025fast,
  title={Fast: Efficient action tokenization for vision-language-action models},
  author={Pertsch, Karl and Stachowicz, Kyle and Ichter, Brian and Driess, Danny and Nair, Suraj and Vuong, Quan and Mees, Oier and Finn, Chelsea and Levine, Sergey},
  journal={arXiv preprint arXiv:2501.09747},
  year={2025}
}

@article{ho2022classifier,
  title={Classifier-free diffusion guidance},
  author={Ho, Jonathan and Salimans, Tim},
  journal={arXiv preprint arXiv:2207.12598},
  year={2022}
}
\bibliographystyle{iclr2027_conference}

\clearpage

\appendix

\section{Handheld demonstrations}
\label{app:umi}

This appendix concerns FastUMI-100K, the handheld data of the first stage. In the second stage, inverse kinematics maps the YUBI demonstrations to the robot instead (Section~\ref{sec:real_robot}).

\paragraph{Frames.}
We express both domains at the flange of the robot arm. DROID records this pose directly. FastUMI-100K records the pose of the handheld device, a position $\mathbf{p}$ and an orientation $R$ decoded from extrinsic $xyz$ Euler angles, in a local frame with $+x$ forward, $+y$ left and $+z$ up on the device. We map it to the common frame by
\begin{align*}
  \mathbf{p}' = \mathbf{p} + R\,\mathbf{b}, \qquad R' = R A, \qquad
  A = \begin{bmatrix} 0 & 0 & 1 \\ 0 & -1 & 0 \\ 1 & 0 & 0 \end{bmatrix}, \qquad
  \mathbf{b} = (-0.25, 0, 0)\,\mathrm{m},
\end{align*}
so that the common $z$ axis points along the approach direction and the reference point lies \qty{25}{cm} behind the recorded pose along it, where it plays the role of the robot flange. The offset is an adopted tool geometry rather than a calibrated one. We verified the axis assignment by correlating pose changes with image motion on 24,618 pairs of adjacent frames from 72 episodes, and all 12 tasks among them selected the same right-handed assignment. Gripper readings are converted to an opening between 0 and 1 in both domains, with the closure recorded by DROID mapped to one minus its value.

\paragraph{Tracking jumps.}
Handheld poses come from visual tracking, which occasionally drops out or jumps. We remove every FastUMI-100K episode with a dropout, two consecutive frames with identical poses, or a jump, a change of the arm speed by more than \qty{500}{mm/s} between consecutive frames. This removes 2,862 of 39,403 episodes. Robot episodes are not filtered, and the input normalization is fitted after the removal.

\section{Speed prior training}
\label{app:training}

\paragraph{Architecture.}
The input sequence is an instruction token followed, for each node, by an image token, a node token and, except after the last node, a query token. The relative pose of the nodes makes windows comparable across domains and setups~\citep{chi2024universal}. Node tokens are linear projections of the seven node features, which are normalized by an affine transform fitted to the 1st and 99th percentiles of the training windows and then rounded in the DCT domain, where the DCT-II coefficients of each feature along the window are rounded to multiples of 1/50~\citep{pertsch2025fast}. Near the ends of an episode, nodes outside the episode are filled with the nearest valid node before rounding. The instruction token is a linear projection of the frozen 1024-dimensional Qwen3-Embedding-0.6B embedding of the instruction at its last token, normalized to unit length~\citep{zhang2025qwen3}. Each image token is a linear projection, after RMSNorm, of the frozen DINOv3 ViT-B/16 features of the wrist-camera frame nearest the node at its native resolution, the class token concatenated with the mean patch token~\citep{simeoni2025dinov3}. A dropped window replaces its 17 image tokens with a learned token. The query tokens are learned embeddings. The transformer has 12 layers of width 384 with six-head bidirectional attention, query-key RMSNorm, rotary position embeddings over consecutive positions and SwiGLU feed-forward layers, and a learned embedding of the domain label conditions the network.
Two linear heads read each query token, one for the arm and one for the gripper.

\paragraph{Optimization.}
In the first stage, windows of both domains are pooled and sampled uniformly with replacement, so each domain contributes in proportion to its number of windows. We train with AdamW at a learning rate of $10^{-4}$ after 1,000 warm-up steps, $\beta = (0.9, 0.95)$, no weight decay, gradient clipping at 1.0 and a batch size of 1,024, and keep an exponential moving average of the weights with decay 0.999. For each domain, 5\% of the episodes are held out for validation, and we use the checkpoint, among those saved every 5,000 updates, whose quantiles best match the held-out speeds. In the second stage, we fine-tune this prior on each YUBI task for at most 3,000 updates at a learning rate of $5 \times 10^{-6}$ and keep the best checkpoint.

\paragraph{Interval durations and clock details.}
We predict speed by classification, which is more robust to noisy targets than squared-error regression~\citep{farebrother2024stop}. Each node takes the time at which the recorded path first reaches its arc length, and the first node of an episode the time at which the arm leaves rest. The two-hot target splits unit mass between the two bin centers nearest to $y_j$, and targets beyond the outermost centers are clipped to them. Quantiles are decoded from the cumulative distribution, interpolating linearly within a bin. With the two speeds independent, the cumulative distribution of the duration $T_j$ is the product of those of $L^{\mathrm{a}}_j / v^{\mathrm{a}}_j$ and $L^{\mathrm{g}}_j / v^{\mathrm{g}}_j$. An actuator that does not move in an interval takes no time. We retime DROID with the FastUMI-100K label at the guidance coefficient $\lambda = 1$ and the YUBI demonstrations at $\lambda = 2$.
The arm travel $L^{\mathrm{a}}_j$ sums $\Delta s_k$ without its gripper term. The gripper travel $L^{\mathrm{g}}_j$ counts \qty{42.5}{mm} per unit of opening in every domain, half the stroke of the Robotiq 2F-85, whereas the arc scale $\rho_{\mathrm{g}}$ only sets how far apart the nodes lie. The speed of the last interval extends over any partial interval at the end of an episode. Episodes with fewer than three nodes keep their recorded timestamps.

\section{Policy training}
\label{app:policy}

\paragraph{Common settings.}
Every policy of Section~\ref{sec:exp} is a $\pi_{0.5}$ policy~\citep{intelligence2025openpi} that predicts chunks of $H = 15$ joint-position actions at \qty{15}{Hz}. Every policy uses train-time RTC~\citep{black2025training} and is conditioned on a prefix of the first actions of the chunk. Our policy and Filtered BC learn chunks that are compressed in time, from retimed or faster demonstrations, so the same inference delay spans more of their actions, and their prefix covers 0 to 5 actions against 0 to 3 for the default policy. Observation frames are selected on the recorded timing, so the frames that a policy sees do not depend on its clock. The joint and gripper dimensions reuse the action normalization of the released checkpoint, and the tempo channels are normalized with statistics computed on the training chunks. Tempo channels are capped at $\log 5$, and a step without motion repeats the channel of the last moving step, or zero before it.

\paragraph{First stage.}
The default policy, Filtered BC and our policy of Section~\ref{sec:robolab} start from the released $\pi_{0.5}$ checkpoint for DROID, whose vision-language backbone stays frozen while the action expert is fine-tuned. Each is trained for up to 20,000 updates with a batch size of 256 and a constant learning rate of $5 \times 10^{-5}$ after 1,000 warm-up steps, and we evaluate one of the checkpoints saved every 5,000 updates. All three are trained on the measured joint positions of DROID rather than the commanded ones, since the commands are tied to the recorded timing.

\paragraph{Second stage.}
The YUBI policies of Section~\ref{sec:real_robot}, the default policy, Filtered BC and ours, all start from our policy of the first stage, so that they differ only in their demonstrations and timing, and all their parameters are fine-tuned. Each is trained for 20,000 updates with a batch size of 32 and a learning rate that decays from $5 \times 10^{-5}$ to $5 \times 10^{-6}$ on a cosine schedule after 1,000 warm-up steps, and we use its last checkpoint. The joint positions are those that inverse kinematics gives for the tracked gripper poses (Section~\ref{sec:real_robot}).

\section{Execution in RoboLab}
\label{app:robolab}

\paragraph{Simulated robot.}
RoboLab's DROID scene carries a Franka Panda, which we replace with the Franka Research 3 of our real robot, whose kinematics are the same. The arm takes the link masses, centers of mass and inertias and the joint position limits of the FR3 model in franka\_description, the firmware velocity limits of Appendix~\ref{app:controller}, which the simulator enforces, and rotor armatures of \qty{0.195}{kg.m^2} on the first four joints and \qty{0.074}{kg.m^2} on the last three. The gripper is Robotiq's PhysX model of the 2F-85, with the finger drive of Robotiq's simulation guide, the closing speed that DROID commands and a pad friction of 0.7.

\paragraph{Playback at a pace.}
The policy predicts $H = 15$ joint waypoints on its base clock, one per control period $\delta = 1/15\,\mathrm{s}$. On the tempo channel of quantile $q$, step $h$ instead lasts $\delta \exp\big(-\tau^{(q)}_h\big)$, and under a uniform speed-up $\alpha$, every step lasts $\delta / \alpha$. Starting from the running reference at the step where the chunk takes over, the waypoints form a timed polyline, which the joint reference follows at the \qty{120}{Hz} physics rate, with each segment's slope as its velocity and zero acceleration. The gripper takes the command of the step in progress.

\paragraph{Tracking.}
A computed-torque controller tracks this reference,
\begin{align*}
  \mathbf{u} = \mathbf{M}(\mathbf{x})\big(k_{\mathrm{p}}(\mathbf{x}_{\mathrm{ref}} - \mathbf{x}) + k_{\mathrm{d}}(\dot{\mathbf{x}}_{\mathrm{ref}} - \dot{\mathbf{x}}) + \ddot{\mathbf{x}}_{\mathrm{ref}}\big) + \mathbf{C}(\mathbf{x}, \dot{\mathbf{x}})\,\dot{\mathbf{x}},
\end{align*}
with $k_{\mathrm{p}} = 300$ and the critically damped $k_{\mathrm{d}} = 2\sqrt{k_{\mathrm{p}}}$, where $\mathbf{M}$ and $\mathbf{C}$ are read from the simulator and gravity is disabled on the arm. The torque is clamped to the effort limits of the simulated arm. The gripper command, an opening between 0 and 1, sets the position target of the gripper. Unlike the controller of Section~\ref{sec:controller}, this execution neither plans within the robot's limits nor joins consecutive chunks with a transition, so the velocity jumps at every waypoint.

\paragraph{Replanning.}
Every policy runs with RTC, conditioned on the unexecuted actions of the running chunk (Appendix~\ref{app:policy}). The policy is queried when the running chunk reaches the waypoint of its open-loop horizon, 8 steps for the default policy and 6 for ours and Filtered BC, and sees the joint state of that step with the last rendered images. Its chunk takes over two steps later, so that its first two actions are the prefix it was conditioned on. The simulation counts this delay in steps rather than in time, so the latency of the policy does not affect the results.

\section{Additional analysis of the speed prior}
\label{app:prior_analysis}

\paragraph{Accuracy.}
The held-out validation episodes hold 363,567 DROID windows from 3,627 episodes and 273,810 FastUMI-100K windows from 1,827 episodes, whose nodes all lie within the episode, and each window contributes the speed of its center interval. Table~\ref{tab:prior_calibration} compares the predicted distributions with the recorded speeds. The predicted median explains 95\% of the variance of held-out log speed on DROID and 93\% on FastUMI-100K, and its pinball loss at the median is a sixth of that of an input-blind predictor on DROID and under a quarter on FastUMI-100K. The prior thus reads most of the pace from the path, the images and the instruction. Its quantiles are close to their nominal levels in both domains, with 89.9\% and 98.7\% of held-out DROID speeds at or below the 0.9 and 0.99 quantiles, and 90.9\% and 99.1\% on FastUMI-100K.

\paragraph{Tracking jitter.}
Handheld poses come from visual tracking, whose jitter differs from that of robot joint encoders. In both domains, the energy that the DCT rounding removes from a window falls with speed (Spearman correlation $-0.47$ on DROID and $-0.40$ on FastUMI-100K), because slow motion leaves relatively more jitter within a window. A prior that sees this residue could read the absence of tracking jitter in robot data as a cue for the fast handheld pace. The same prior trained without the rounding indeed transfers more speed, raising the median speed of DROID by $1.56\times$ rather than $1.28\times$ when read with the FastUMI-100K label, and its retimed DROID episodes are less smooth, with a SPARC of $-3.45$ rather than $-3.09$ under the median clock. It is also less robust to tracking noise: Gaussian jitter of \qty{0.9}{mm} and \ang{0.35} per axis added to the DROID windows lowers its predicted median speed by 63\% and raises its speed-up to $2.04\times$, against 38\% and $1.48\times$ with the rounding. With the rounding, the residual of the prior is uncorrelated with the removed energy (Spearman correlation $0.008$ on DROID and $0.001$ on FastUMI-100K). The rounding thus makes the prior more robust to tracking noise, and the transferred pace mainly reflects how handheld demonstrators move rather than how they are tracked.

\paragraph{Retimed DROID demonstrations.}
Reading DROID with the FastUMI-100K label raises the predicted speed by $1.28\times$ at the median, $1.38\times$ at the 0.9 quantile and $1.54\times$ at the 0.99 quantile on average. On 2,000 held-out DROID episodes, the median clock shortens the episodes by a factor of 1.54 and the clock of the 0.99 quantile by a factor of 2.66. Their SPARC is $-3.09$ and $-3.07$, against $-3.91$ for the recorded motion and $-3.87$ and $-3.74$ under uniform acceleration to the same durations.

\begin{table}[t]
\centering
\caption{Accuracy of the speed prior on held-out validation windows, each domain read with its own label. Coverage is the fraction of recorded speeds at or below the predicted quantile. Pinball losses are in natural log speed, for the prior and for an input-blind predictor that returns the training quantiles of the domain. $R^2$ is the fraction of the variance of recorded log speed explained by the predicted median, within episodes after removing each episode's mean and between episodes among the episode means, over the episodes with at least five windows. The variance between episodes is 18\% of the total on DROID and 12\% on FastUMI-100K, and the $R^2$ over all windows, 0.95 and 0.93, is the average of the two columns weighted by these shares.}
\label{tab:prior_calibration}
\resizebox{\columnwidth}{!}{%
\begin{tabular}{@{}lcccccccccc@{}}
  \toprule
  & \multicolumn{4}{c}{Coverage (\%)} & \multicolumn{2}{c}{Pinball, $q = 0.5$} & \multicolumn{2}{c}{Pinball, $q = 0.9$} & \multicolumn{2}{c}{$R^2$ of the median} \\
  \cmidrule(l){2-5} \cmidrule(l){6-7} \cmidrule(l){8-9} \cmidrule(l){10-11}
  Domain & $q = 0.1$ & $0.5$ & $0.9$ & $0.99$ & prior & input-blind & prior & input-blind & within & between episodes \\
  \midrule
  DROID        & 10.3 & 50.0 & 89.9 & 98.7 & 0.050 & 0.294 & 0.021 & 0.098 & 0.94 & 0.97 \\
  FastUMI-100K &  9.0 & 49.6 & 90.9 & 99.1 & 0.045 & 0.204 & 0.020 & 0.077 & 0.92 & 0.98 \\
  \bottomrule
\end{tabular}}
\end{table}

\begin{table}[t]
\centering
\caption{Speed around gripper events on held-out validation windows. Windows are grouped by the arc length from their center interval to the nearest interval in which the gripper opening changes by more than 0.05. Speeds are medians relative to the transit group, whose recorded and predicted medians are \qtylist{228;228}{mm/s} on DROID and \qtylist{274;273}{mm/s} on FastUMI-100K. In the same interval the gripper moves and keeps its source pace. The median clock reads DROID with the FastUMI-100K label, and its speed-up is the geometric mean ratio of its median to the median read with the DROID label.}
\label{tab:prior_phase}
\resizebox{\columnwidth}{!}{%
\begin{tabular}{@{}lccccccccc@{}}
  \toprule
  & \multicolumn{5}{c}{DROID} & \multicolumn{3}{c}{FastUMI-100K} \\
  \cmidrule(l){2-6} \cmidrule(l){7-9}
  Distance to gripper event & Share (\%) & Recorded & Prior & Median clock & Speed-up & Share (\%) & Recorded & Prior \\
  \midrule
  Same interval         & 16.6 & 0.24 & 0.24 & 0.23 & 1.08 & 14.9 & 0.58 & 0.58 \\
  \qtyrange{2}{4}{cm}   &  9.1 & 0.45 & 0.44 & 0.57 & 1.57 & 10.5 & 0.67 & 0.67 \\
  \qtyrange{6}{10}{cm}  & 12.1 & 0.68 & 0.68 & 0.76 & 1.39 & 13.4 & 0.84 & 0.84 \\
  \qtyrange{12}{16}{cm} &  8.9 & 0.84 & 0.83 & 0.88 & 1.32 & 10.2 & 0.98 & 0.98 \\
  Transit, none within \qty{16}{cm} & 53.3 & 1.00 & 1.00 & 1.00 & 1.28 & 50.9 & 1.00 & 1.00 \\
  \bottomrule
\end{tabular}}
\end{table}

\section{Controller}
\label{app:controller}

Algorithm~\ref{alg:controller} summarizes one cycle of the controller. Inequalities between vectors hold per joint, and $\oslash$ denotes elementwise division. $\mathbf{M}$, $\mathbf{C}$ and $\mathbf{g}$ are the mass matrix, the Coriolis matrix and the gravity torque of the arm with its tool, $\bar{\mathbf{v}}(\mathbf{x})$ is the robot's position-dependent velocity bound, $\bar{\mathbf{u}}$ and $\bar{\dot{\mathbf{u}}}$ are the torque and torque rate limits, and the acceleration scale $\tilde{\mathbf{a}} = (\bar{\mathbf{u}} - |\mathbf{g}(\mathbf{x}_b)|) \oslash \operatorname{diag}\mathbf{M}(\mathbf{x}_b)$ is the acceleration that the torque limit leaves after gravity. The planning margins $\beta_v, \beta_a, \beta_u \le 1$ scale the velocity bound, the acceleration scale and the torque limits.

\begin{algorithm}[t]
\caption{Pace aware chunk execution}
\label{alg:controller}
\small
\begin{algorithmic}[1]
\Require policy $\pi$, tempo channel $q$ or speed-up $\alpha$, robot limits $\mathcal{L}$, box radius $\eta$, blend tolerance $\epsilon$
\While{the task runs} \Comment{the robot tracks the running plan $P$}
  \State $o \gets \Call{Observe}{P}$ \Comment{at $t_o$; with RTC, $o$ includes the unexecuted actions of $P$}
  \State $(\mathbf{w}_h, c_h, \tau^{(q)}_h)_{h=1}^{H} \gets \pi(o)$, \ $r_h \gets \exp\big(\tau^{(q)}_h\big)$ or $r_h \gets \alpha$ \Comment{$t_{\mathrm{vla}}$}
  \State $(t_b, \mathbf{s}_b, h_0, t^\star) \gets \Call{Reserve}{P}$ \Comment{$h_0 = \lfloor p(t_b) \rfloor + 1$, $p(t^\star) = h_0$}
  \State $\bm{\xi} \gets \Call{Draft}{\mathbf{w}_{h_0:H}, c_{h_0:H}, r_{h_0:H}}$ \Comment{$t_{\mathrm{solve}}$ until \textsc{Install}}
  \State $\bar{\mathbf{x}} \gets \Call{Splice}{\mathbf{s}_b, t^\star, \bm{\xi}}$
  \State $P' \gets \Call{Smooth}{\mathbf{s}_b, \bar{\mathbf{x}}}$
  \If{a step fails or not \Call{Verify}{$P', \mathcal{L}$}}
    \State $P' \gets \Call{Brake}{\mathbf{s}_b}$
  \EndIf
  \State \Call{Install}{$P', t_b$} \Comment{$t_{\mathrm{install}}$, then $P \gets P'$ at $t_b$}
\EndWhile
\Statex
\Function{Draft}{$\mathbf{w}, c, r$}
  \State $\bm{\gamma} \gets \Call{MinJerkPath}{\mathbf{w}, \eta}$ \Comment{geometry on the base clock $\sigma$}
  \State \Return $\Call{TimeOptimal}{\bm{\gamma}, \mathcal{L}, T^{\min}(c, r)}$ \Comment{timing, each step no shorter than $T^{\min}_h$}
\EndFunction
\Statex
\Function{Splice}{$\mathbf{s}_b, t^\star, \bm{\xi}$}
  \State start $\bm{\xi}$ at $t^\star$ \Comment{$\bm{\xi}(t)$ is due at $t$}
  \For{each entry time $t_e \ge t^\star$} \Comment{$\bm{\xi}(t^\star) = \mathbf{w}_{h_0}$}
    \State $\bm{\psi}_e \gets \Call{Ruckig}{\mathbf{s}_b, \bm{\xi}(t_e), \mathcal{L}}$ \Comment{fastest transition, duration $D_e$}
    \State $\ell_e \gets \max\big(0,\ t_b + D_e - t_e\big)$ \Comment{lateness}
  \EndFor
  \State $e^\star \gets \operatorname*{arg\,lexmin}_e\,(\ell_e, b_e, e)$ \Comment{$b_e$: excess over $\epsilon$ at the skipped waypoints}
  \State \Return $\bm{\psi}_{e^\star} \oplus \bm{\xi}|_{t \ge t_{e^\star}}$ \Comment{$\oplus$ concatenates in time; $\bm{\psi}$ to $\mathbf{w}_{h_0}$ ends no earlier than $t^\star$}
\EndFunction
\Statex
\Function{Smooth}{$\mathbf{s}_b, \bar{\mathbf{x}}$}
  \State $P' \gets \Call{JerkQP}{\mathbf{s}_b, \bar{\mathbf{x}}, \eta, \mathcal{L}}$ \Comment{boxes at the due times $t_h$ on $\bar{\mathbf{x}}$}
  \If{$P'$ is infeasible or not \Call{Verify}{$P', \mathcal{L}$}}
    \State $P' \gets$ the same program with the boxes as a penalty
  \EndIf
  \State \Return $P'$
\EndFunction
\end{algorithmic}
\end{algorithm}

\paragraph{Timing.}
A worker runs inference and the solver, a \qty{100}{Hz} client schedules observations and installs plans, and a \qty{1}{kHz} torque executor plays the active plan, sampled every \qty{1}{ms} with its progress $p$, and swaps in the new plan exactly at $t_b$. Each chunk runs open loop for the horizon of Table~\ref{tab:setup_settings}, and the next observation is taken when $P$ reaches the first whole step at most one pipeline latency, the sum of the 90th percentiles of $t_{\mathrm{vla}}$, $t_{\mathrm{solve}}$ and $t_{\mathrm{install}}$, before that step, so that the observation lies on the action grid of $P$. The switch time $t_b$ lies the 90th percentiles of $t_{\mathrm{solve}}$ and $t_{\mathrm{install}}$ plus one client period ahead. With train-time RTC~\citep{black2025training}, the policy keeps the first $d$ unexecuted actions of $P$ fixed, where $d$ is the progress of $P$ over the 95th percentile of the time from an observation to the installation of its plan, rounded up and clipped to the delays that the policy was trained with, and $t_b$ is delayed until $P$ reaches progress $d$. These delays are $\{0, \dots, 5\}$ for our policy and Filtered BC, whose chunks are compressed in time so that the same latency spans more of their steps, and $\{0, \dots, 3\}$ for the default policy (Appendix~\ref{app:policy}). The client drops the first $\lfloor p(t_b) \rfloor$ actions, exactly the fixed ones when $t_b$ is met as planned, and passes the consumed fraction $\phi = p(t_b) - \lfloor p(t_b) \rfloor$ of step $h_0$.

\paragraph{Target schedule.}
Step $h > h_0$ lasts at least $T^{\min}_h = \max\big(\delta / r_h,\ |c_h - c_{h-1}| / \mu\big)$, where $\mu$ is the gripper speed, so that the arm does not outrun the gripper, and step $h_0$ at least $T^{\min}_{h_0} = \max\big((1 - \phi)\,\delta / r_{h_0},\ |c_{h_0} - c_b| / \mu\big)$, with $c_b$ the gripper command at $t_b$. The due time $t^\star$ is likewise postponed until the gripper can reach $c_{h_0}$. When $P$ ends before step $h_0$, as for the first chunk or after a braking plan, $t^\star = t_b + T^{\min}_{h_0}$.

\paragraph{\textsc{Draft}.}
The path $\bm{\gamma}$ is piecewise quadratic in $\sigma$, with the acceleration $\mathbf{a}_m$ constant over rows of \qty{32}{ms}. It starts at $\mathbf{w}_{h_0}$ with the secant velocity toward $\mathbf{w}_{h_0+1}$ and minimizes $\sum_m \|(\mathbf{a}_m - \mathbf{a}_{m-1}) \oslash \tilde{\mathbf{a}}\|^2$ subject to $\|\bm{\gamma}(h\delta) - \mathbf{w}_h\|_\infty \le \eta = \qty{5}{mrad}$ and the joint range. This quadratic program has no velocity, acceleration or torque rows, so the polyline through the waypoints is feasible, and we solve it with HPIPM~\citep{frison2020hpipm} and fall back to that polyline if the solver fails. We fit a cubic spline to $\bm{\gamma}$ with knots every \qty{24}{ms} of $\sigma$ and retime it with TOPP-RA~\citep{pham2018new} on 64 gridpoints plus the knots, under the scaled velocity, acceleration and torque limits, the torque from the full inverse dynamics, and under $\dot{\sigma} \le 1.01\,\bar{r}_\nu$ on each knot span $\nu$, where $\bar{r}_\nu$ is the length of the span divided by its minimum duration under $T^{\min}_h$. The entry rate is $0.99$ times the largest feasible one, and the terminal rate is not forced to zero, so a plan ends moving and the next plan takes over before its end.

\paragraph{\textsc{Splice}.}
Candidate entries are spaced \qty{16}{ms} apart from $t_b$ up to step 8, with $\bm{\xi}(t_e) = \mathbf{w}_{h_0}$ for $t_e \le t^\star$. A transition to $\mathbf{w}_{h_0}$ is stretched so that it arrives no earlier than $t^\star$, unless it would then turn back, and a postponement of $t^\star$ for the gripper slows the first span of $\bm{\xi}$ instead. Ruckig~\citep{berscheid2021jerk} bounds each transition by $\beta_v \bar{\mathbf{v}}$, by $\beta_a \tilde{\mathbf{a}}$ and by the jerk $\beta_u \bar{\dot{\mathbf{u}}} \oslash \operatorname{diag}\mathbf{M}(\mathbf{x}_b)$ that the torque rate limit allows. The selected entry is
\begin{align*}
  e^\star = \operatorname*{arg\,lexmin}_{e}\, \big(\ell_e,\, b_e,\, e\big), \qquad
  b_e = \max\Big(0,\, \max_{h \in \mathcal{S}_e}\, \min_{t}\, \|\bm{\psi}_e(t) - \mathbf{w}_h\|_\infty / \epsilon - 1\Big),
\end{align*}
where $\mathcal{S}_e$ holds the waypoints that the entry skips and $\epsilon = \qty{0.05}{rad}$. The rest of $\bm{\xi}$ follows the transition without a gap, and the gripper commands of skipped waypoints are issued by the end of the transition.

\paragraph{\textsc{Smooth}.}
The quadratic program holds the jerk $\bm{\zeta}_i$ constant over rows of $T_{\mathrm{r}} = \qty{16}{ms}$, starts from $\mathbf{s}_b$ and minimizes $\sum_i \|T_{\mathrm{r}}\,\bm{\zeta}_i \oslash \tilde{\mathbf{a}}\|^2$ without a tracking term. At the end of every row it enforces the joint range, the scaled velocity and acceleration limits, the scaled torque limit with the dynamics linearized along $\bar{\mathbf{x}}$, and the torque rate limit $|\mathbf{M}(\bar{\mathbf{x}}_i)\,\bm{\zeta}_i| \le \beta_u \bar{\dot{\mathbf{u}}}$. Each waypoint after the entry must lie within $\eta$ at $t_h$, and the terminal velocity within \qty{0.05}{rad/s} of that of $\bm{\xi}$. If the program is infeasible or its plan fails verification, the boxes are replaced by a linear penalty of $10^3$ per radian.

\paragraph{Verification and fallbacks.}
Each plan is checked against the unscaled limits, the velocity, acceleration and joint range every \qty{1}{ms} and the torque from the full inverse dynamics and its rate every \qty{8}{ms}, and is accepted if it exceeds no limit by more than 5\% and no joint range by more than \qty{1}{mrad}.
Otherwise, and also when TOPP-RA cannot time the path or Ruckig finds no transition, the solver returns a braking plan, the program of \textsc{Smooth} without boxes and ending at rest. Its horizon is Ruckig's time-optimal stopping time rounded up to whole rows and at least \qty{48}{ms}, and is doubled up to twice when the program is infeasible or its plan fails verification. The solver refuses a chunk only when every braking plan fails or the solve takes longer than \qty{1}{s}. After a refusal or a plan that arrives after $t_b$, the client keeps $P$ and solves the remaining actions again from a new switch time, at least \qty{33}{ms} after the previous request, and once $P$ ends it drops the chunk and takes a new observation. When $P$ is about to end in motion with no plan pending, the client requests a braking plan itself, and past the end of any plan the executor brakes every joint at \qty[per-mode = symbol]{10}{\radian\per\second\squared}.

\paragraph{Low-level execution.}
The executor applies at \qty{1}{kHz} the joint torque
\begin{align*}
  \mathbf{u} = \big(\mathbf{J}^\top\mathbf{K}^{\mathrm{c}}_{\mathrm{p}}\mathbf{J} + \mathbf{K}_{\mathrm{p}}\big)(\mathbf{x}_{\mathrm{ref}} - \mathbf{x})
  + \big(\mathbf{J}^\top\mathbf{K}^{\mathrm{c}}_{\mathrm{d}}\mathbf{J} + \mathbf{K}_{\mathrm{d}}\big)(\dot{\mathbf{x}}_{\mathrm{ref}} - \dot{\mathbf{x}}) + \mathbf{C}(\mathbf{x}, \dot{\mathbf{x}})\,\dot{\mathbf{x}} + \kappa\,\mathbf{M}(\mathbf{x})\,\ddot{\mathbf{x}}_{\mathrm{ref}},
\end{align*}
clamped to the torque limits, where $(\mathbf{x}_{\mathrm{ref}}, \dot{\mathbf{x}}_{\mathrm{ref}}, \ddot{\mathbf{x}}_{\mathrm{ref}})$ is the current plan sample, $\mathbf{J}$ is the end-effector Jacobian, $\mathbf{K}_{\mathrm{p}}, \mathbf{K}_{\mathrm{d}}$ and $\mathbf{K}^{\mathrm{c}}_{\mathrm{p}}, \mathbf{K}^{\mathrm{c}}_{\mathrm{d}}$ are the joint and Cartesian stiffness and damping gains, and the robot compensates gravity. The default gains of DROID's Polymetis configuration~\citep{khazatsky2024droid,lin2021polymetis} are $\mathbf{K}_{\mathrm{p}} = \operatorname{diag}(40, 30, 50, 25, 35, 25, 10)$, $\mathbf{K}_{\mathrm{d}} = \operatorname{diag}(4, 6, 5, 5, 3, 2, 1)$, $\mathbf{K}^{\mathrm{c}}_{\mathrm{p}} = \operatorname{diag}(400, 400, 400, 15, 15, 15)$ and $\mathbf{K}^{\mathrm{c}}_{\mathrm{d}} = \operatorname{diag}(37, 37, 37, 2, 2, 2)$ in SI units. Our controller scales the stiffness gains by 4 and the damping gains by 2 and uses $\kappa = 1$. Gripper commands are sent at \qty{30}{Hz}.

\paragraph{PD baseline.}
Following the openpi DROID example~\citep{khazatsky2024droid,intelligence2025openpi}, the PD baseline holds each waypoint as a position target for $T^{\min}_h$, a \qty{15}{Hz} stepped reference at $\alpha = 1$, and caps each new target at \qty{0.2}{rad} from the measured position, as DROID's joint velocity interface does. It uses the torque law above with the default gains, $\dot{\mathbf{x}}_{\mathrm{ref}} = \ddot{\mathbf{x}}_{\mathrm{ref}} = \mathbf{0}$ and $\kappa = 0$, the hybrid joint impedance controller of Polymetis~\citep{lin2021polymetis}, and shares our observation timing, switch times and RTC. We keep the default gains and the cap because the stepped reference is unsafe otherwise. Without the cap it triggered the robot's safety reflex in four of five trash cleaning episodes, and with our stiffer gains it did so even at half the pace.

\paragraph{Settings.}
Table~\ref{tab:setup_settings} lists the settings that differ between the two setups, and Table~\ref{tab:controller_settings} those of the two controllers. Our controller plans within the firmware limits of the Franka Research 3, with velocity maxima $(2.62, 2.62, 2.62, 2.62, 5.26, 4.18, 5.26)\,\mathrm{rad/s}$, torque limits $(87, 87, 87, 87, 12, 12, 12)\,\mathrm{Nm}$ and a torque rate limit of \qty{1000}{Nm/s}. 
On YUBI, inference and the solve round trip each take \qty{28}{ms} at the median.

\begin{table}[t]
\centering
\small
\caption{Settings that differ between the setups. The DROID zero-shot tasks of Appendix~\ref{app:zero-shot-eval} are placing bananas and placing cubes.}
\label{tab:setup_settings}
\begin{tabular}{@{}lll@{}}
\toprule
Setting & YUBI & DROID zero-shot \\
\midrule
Gripper & Yubi DYNAMIXEL & Robotiq \\
Open-loop horizon & 8 steps & 6 for ours, 8 others \\
Executed tempo quantiles & 0.9, 0.95 and 0.99 & 0.99 \\
\bottomrule
\end{tabular}
\end{table}

\begin{table}[t]
\centering
\small
\caption{Settings of our controller (Ours) and the PD baseline (PD), shared by both setups. An entry that spans both columns applies to each of them.}
\label{tab:controller_settings}
\begin{tabular}{@{}lll@{}}
\toprule
Setting & Ours & PD \\
\midrule
Chunk & \multicolumn{2}{l}{$H = 15$ actions, $\delta = 1/15\,\mathrm{s}$} \\
Stiffness and damping gains, feedforward $\kappa$ & $4\times$, $2\times$, 1 & $1\times$, $1\times$, 0 \\
Step cap & none & \qty{0.2}{rad} \\
\bottomrule
\end{tabular}
\end{table}

\section{Real-robot evaluation}
\label{app:real_robot}

Each condition runs 30 episodes on stacking cups and folding towel and 5 episodes on cleaning trash. The time limit $t_{\max}$ is \qty{10}{s} for cups and towel and \qty{60}{s} for trash. The completion time $t_i$ of a successful episode $i$ is read from the measured gripper. It is the time at which the release that places the second cup has progressed by 90\% in stacking cups, and the time at which its last release has progressed by 90\% in folding towel.

\paragraph{Success rate and time.}
SR is the fraction of successful episodes, and ATR is the mean of $t_i$ over the successful episodes.

\paragraph{Throughput with regret.}
TPR~\citep{arachchige2025sail} rewards each success by the inverse of its completion time and penalizes each failure by the inverse of the time limit. We report it in successes per minute, 60 times the per-second definition of \citet{arachchige2025sail},
\begin{align*}
  \mathrm{TPR} = \frac{60}{N} \sum_{i=1}^{N} \Big( \frac{S_i}{t_i} - \frac{1 - S_i}{t_{\max}} \Big),
\end{align*}
where $N$ is the number of episodes, $S_i = 1$ if episode $i$ succeeds and $S_i = 0$ otherwise, and times are in seconds.

\paragraph{Throughput.}

In cleaning trash, episode $i$ places $K_i$ of the 15 trash items in the bin, and we define its duration $T_i$ as the time at which the last item is placed if all 15 are placed, and $t_{\max} = \qty{60}{s}$ otherwise. We report throughput in items per hour (UPH), computed over all $N$ episodes as
\[
  \mathrm{UPH} = 3600 \, \frac{\sum_{i=1}^{N} K_i}{\sum_{i=1}^{N} T_i}.
\]

\paragraph{Smoothness.}
SPARC measures the smoothness of the end-effector speed~\citep{balasubramanian2015analysis}, where higher values are smoother. We resample the measured flange position of each episode, saved in its step log, at \qty{15}{Hz} by linear interpolation from its first sample, take its speed $v(t) = \|\dot{\mathbf{p}}(t)\|$ by central differences and compute the arc length of the normalized magnitude spectrum $A(f)$ of the speed, zero-padded to four times the next power of two of its length,
\begin{align*}
  \mathrm{SPARC} = -\sum_{f_k \le f_*} \sqrt{\Big(\frac{\Delta f_k}{f_*}\Big)^2 + \big(\Delta A_k\big)^2},
\end{align*}
where the cutoff $f_*$ is set by an amplitude threshold of 0.05 and a maximum of \qty{10}{Hz}, which the \qty{15}{Hz} sampling limits to its Nyquist frequency of \qty{7.5}{Hz}. A successful episode is evaluated up to its completion and a failed one up to the end of its recording. We report the mean over all episodes on which SPARC can be computed, including the failed ones.

\section{Zero-shot evaluation}
\label{app:zero-shot-eval}

\paragraph{Setup.}
We evaluate the default policy and our policy of Section~\ref{sec:robolab}, which are fine-tuned only on DROID, on the Franka Research 3 of Section~\ref{sec:real_robot} fitted with the Robotiq 2F-85 gripper of DROID, without further fine-tuning, on two tasks. The robot has \qty{30}{s} to put two bananas into a tray and \qty{60}{s} to put nine gray cubes into a white bin (Figure~\ref{fig:real_world_tasks}). On both tasks, the default policy runs under the PD baseline and under our controller at $\alpha = 1$, and our policy runs under our controller at $\alpha = 1$ and on its 0.99 tempo channel. Table~\ref{tab:setup_settings} lists the other settings.

\paragraph{Metrics.}
We run 30 episodes per condition. The operator marks each item with a key press as it lands in the tray or bin.
The first cube episode of the default policy under the PD baseline ran away and was aborted, so that condition keeps the 30 episodes that follow it.
An episode succeeds if it places all $n$ items within $t_{\max}$, with $n = 2$ and $t_{\max} = \qty{30}{s}$ for bananas and $n = 9$ and $t_{\max} = \qty{60}{s}$ for cubes. UPH pools the $N = 30$ episodes of a condition,
\begin{align*}
  \mathrm{UPH} = \frac{3600 \sum_{i=1}^{N} K_i}{\sum_{i=1}^{N} T_i},
\end{align*}
where $K_i$ is the number of items that episode $i$ places within $t_{\max}$, and $T_i$ is the time at which it places the last one if $K_i = n$ and $t_{\max}$ otherwise. SPARC is computed as in Appendix~\ref{app:real_robot}, from the start of the policy to the last item, or to $t_{\max}$ if the episode does not place all items. SPARC uses the step logs of all 30 episodes, including the cube episode of our policy on the 0.99 tempo channel whose separate controller log was overwritten after a pause.

\paragraph{Results.}
Table~\ref{tab:banana-eval} and Figure~\ref{fig:zero-shot-eval} report the results on bananas. On its 0.99 tempo channel, our policy places both bananas in all 30 episodes, at a UPH of 492.9. Under our controller, the default policy places both in 22 episodes ($p = 0.003$, two-sided Mann-Whitney U test on the number of bananas), at a UPH of 231.7, and under the PD baseline in 17, at 200.7. On its base clock, our policy places both in 29 episodes at a UPH of 376.7. It also moves most smoothly there, with a SPARC of $-4.13$ against $-4.99$ for the default policy under our controller ($p < 0.001$), and on the 0.99 tempo channel its SPARC of $-4.53$ remains above that of the default policy ($p = 0.025$).

Table~\ref{tab:cube-eval} and Figure~\ref{fig:zero-shot-cube} report the results on cubes. On its 0.99 tempo channel, our policy places 8.27 cubes on average and all nine in 12 of 30 episodes. Under our controller, the default policy places 5.90 and never all nine ($p < 0.001$), so UPH rises from 354.0 to 511.1. On its base clock, our policy places 7.43 cubes at a UPH of 446.7, and the 0.99 tempo channel raises this to 8.27 ($p = 0.001$). Under the PD baseline, the default policy places only 3.90 cubes, at a UPH of 234.0. Our policy moves most smoothly on its base clock, with a SPARC of $-5.46$ against $-6.26$ for the default policy under our controller ($p < 0.001$). On the 0.99 tempo channel, its SPARC of $-6.18$ is close to that of the default policy ($p = 0.63$).

\begin{table}[hbt!]
    \centering
    \setlength{\tabcolsep}{3.5pt}
    \caption{Real-robot results on the placing bananas task (``put bananas into the tray''), with 30 trials per setting. PD is the baseline controller of Appendix~\ref{app:controller}, Ours (1x) runs our controller at $\alpha = 1$ and Ours (q99) on the tempo channel of the 0.99 quantile. Bananas is the mean number of bananas placed within \qty{30}{s}, and Both the number of trials that place both.}
    \label{tab:banana-eval}
    \begin{tabular}{llcccc}
    \toprule
             &             & \multicolumn{4}{c}{Placing bananas} \\
             \cmidrule(l){3-6}
       Model & Controller  & Bananas$\uparrow$ & Both$\uparrow$ & UPH$\uparrow$ & SPARC$\uparrow$ \\
       \midrule
       Default & PD          & 1.57 & 17 & 200.7 & \textminus5.28 \\
       Default & Ours (1x)   & 1.67 & 22 & 231.7 & \textminus4.99 \\
       \midrule
       Ours    & Ours (1x)   & 1.97 & 29 & 376.7 & \textbf{\textminus4.13} \\
       Ours    & Ours (q99)  & \textbf{2.00} & \textbf{30} & \textbf{492.9} & \textminus4.53 \\
    \bottomrule
    \end{tabular}
\end{table}

\begin{table}[hbt!]
    \centering
    \setlength{\tabcolsep}{3.5pt}
    \caption{Real-robot results on the placing cubes task (``put all the gray cubes into the white bin''), with 30 trials per setting. PD is the baseline controller of Appendix~\ref{app:controller}, Ours (1x) runs our controller at $\alpha = 1$ and Ours (q99) on the tempo channel of the 0.99 quantile. Cubes is the mean number of cubes placed within \qty{60}{s}, and All 9 the number of trials that place all nine.}
    \label{tab:cube-eval}
    \begin{tabular}{llcccc}
    \toprule
             &             & \multicolumn{4}{c}{Placing cubes} \\
             \cmidrule(l){3-6}
       Model & Controller  & Cubes$\uparrow$ & All 9$\uparrow$ & UPH$\uparrow$ & SPARC$\uparrow$ \\
       \midrule
       Default & PD          & 3.90 & 0 & 234.0 & \textminus8.19 \\
       Default & Ours (1x)   & 5.90 & 0 & 354.0 & \textminus6.26 \\
       \midrule
       Ours    & Ours (1x)   & 7.43 & 2 & 446.7 & \textbf{\textminus5.46} \\
       Ours    & Ours (q99)  & \textbf{8.27} & \textbf{12} & \textbf{511.1} & \textminus6.18 \\
    \bottomrule
    \end{tabular}
\end{table}

\definecolor{cpd}{HTML}{9E9E9E}
\definecolor{cdef}{HTML}{3C3C3A}
\definecolor{coursb}{HTML}{6A93D4}
\definecolor{cours}{HTML}{2A5DB0}
\begin{figure}[t]
\centering
\begin{tikzpicture}
\begin{axis}[
  width=0.62\linewidth, height=0.40\linewidth,
  xmin=0, xmax=30, ymin=0, ymax=102,
  xtick={0,5,...,30}, ytick={0,25,50,75,100},
  xlabel={Time (s)}, ylabel={Cumulative progress (\%)},
  ymajorgrids, grid style={black!8, line width=0.4pt},
  axis lines*=left, axis line style={black!55, line width=0.5pt},
  tick style={black!55}, tick align=outside, major tick length=2.5pt,
  tick label style={font=\footnotesize, text=black!70},
  label style={font=\small},
  every axis plot/.append style={const plot, line join=round, line cap=round},
  legend style={draw=none, fill=white, fill opacity=0.92, text opacity=1,
    font=\footnotesize, at={(1.04,0.5)}, anchor=west,
    row sep=1pt, inner sep=3pt},
  legend image post style={xscale=1.2},
  legend cell align=left, reverse legend, clip=false,
]
\addplot[cpd, line width=1pt, dash pattern=on 3pt off 1.8pt]
  coordinates {
    (0.00,0.00) (7.70,1.67) (7.76,3.33) (7.95,5.00) (8.08,6.67) (8.38,8.33) (8.73,10.00) (8.82,11.67)
    (8.87,13.33) (9.03,15.00) (9.04,16.67) (9.15,18.33) (9.15,20.00) (9.21,21.67) (9.30,23.33) (9.69,25.00)
    (9.90,26.67) (10.80,28.33) (10.84,30.00) (12.23,31.67) (12.62,33.33) (13.06,35.00) (13.41,36.67) (14.48,38.33)
    (16.01,40.00) (16.87,41.67) (16.99,43.33) (17.61,45.00) (18.09,46.67) (20.12,48.33) (20.28,50.00) (23.88,51.67)
    (24.36,53.33) (24.98,55.00) (25.08,56.67) (25.22,58.33) (25.75,60.00) (25.91,61.67) (26.08,63.33) (26.34,65.00)
    (27.05,66.67) (27.12,68.33) (27.84,70.00) (28.35,71.67) (28.54,73.33) (28.66,75.00) (29.04,76.67) (29.07,78.33)
    (30.00,78.33)};
\addlegendentry{Default / PD}
\addplot[only marks, mark=*, mark size=1.3pt, cpd, forget plot] coordinates {(29.07,78.33)};
\addplot[cdef, line width=1pt]
  coordinates {
    (0.00,0.00) (5.63,1.67) (5.96,3.33) (6.42,5.00) (6.47,6.67) (6.51,8.33) (6.54,10.00) (6.54,11.67)
    (6.57,13.33) (6.57,15.00) (6.67,16.67) (6.99,18.33) (7.19,20.00) (7.30,21.67) (7.46,23.33) (7.47,25.00)
    (7.47,26.67) (7.49,28.33) (7.49,30.00) (7.92,31.67) (8.02,33.33) (8.35,35.00) (9.79,36.67) (10.44,38.33)
    (11.20,40.00) (11.53,41.67) (11.88,43.33) (13.18,45.00) (17.51,46.67) (18.91,48.33) (19.04,50.00) (19.14,51.67)
    (19.87,53.33) (20.07,55.00) (21.34,56.67) (22.46,58.33) (23.62,60.00) (24.14,61.67) (24.37,63.33) (24.48,65.00)
    (24.96,66.67) (25.46,68.33) (25.62,70.00) (26.28,71.67) (26.29,73.33) (26.68,75.00) (28.27,76.67) (28.77,78.33)
    (29.13,80.00) (29.65,81.67) (30.00,83.33) (30.00,83.33)};
\addlegendentry{Default / Ours (1x)}
\addplot[only marks, mark=*, mark size=1.3pt, cdef, forget plot] coordinates {(30.00,83.33)};
\addplot[coursb, line width=1pt]
  coordinates {
    (0.00,0.00) (4.98,1.67) (5.02,3.33) (5.07,5.00) (5.09,6.67) (5.17,8.33) (5.35,10.00) (5.41,11.67)
    (5.48,13.33) (5.52,15.00) (5.58,16.67) (5.60,18.33) (5.67,20.00) (5.67,21.67) (5.68,23.33) (5.76,25.00)
    (5.78,26.67) (5.86,28.33) (5.88,30.00) (5.96,31.67) (6.01,33.33) (6.02,35.00) (6.05,36.67) (6.17,38.33)
    (6.22,40.00) (6.50,41.67) (6.55,43.33) (6.66,45.00) (8.21,46.67) (9.22,48.33) (13.88,50.00) (14.77,51.67)
    (14.99,53.33) (15.60,55.00) (16.10,56.67) (16.28,58.33) (16.28,60.00) (16.39,61.67) (16.41,63.33) (16.56,65.00)
    (16.89,66.67) (16.90,68.33) (16.97,70.00) (17.49,71.67) (17.91,73.33) (17.91,75.00) (18.16,76.67) (18.47,78.33)
    (18.88,80.00) (19.10,81.67) (20.41,83.33) (20.58,85.00) (20.84,86.67) (21.25,88.33) (21.27,90.00) (21.46,91.67)
    (21.55,93.33) (22.59,95.00) (23.42,96.67) (25.44,98.33) (30.00,98.33)};
\addlegendentry{Ours / Ours (1x)}
\addplot[only marks, mark=*, mark size=1.3pt, coursb, forget plot] coordinates {(25.44,98.33)};
\addplot[cours, line width=1.5pt]
  coordinates {
    (0.00,0.00) (3.66,1.67) (3.84,3.33) (3.88,5.00) (3.88,6.67) (3.91,8.33) (4.03,10.00) (4.03,11.67)
    (4.05,13.33) (4.06,15.00) (4.06,16.67) (4.06,18.33) (4.11,20.00) (4.14,21.67) (4.19,23.33) (4.25,25.00)
    (4.26,26.67) (4.31,28.33) (4.39,30.00) (4.54,31.67) (4.54,33.33) (4.55,35.00) (4.58,36.67) (4.60,38.33)
    (4.80,40.00) (4.82,41.67) (4.89,43.33) (4.95,45.00) (5.08,46.67) (5.15,48.33) (8.26,50.00) (11.48,51.67)
    (11.49,53.33) (11.62,55.00) (11.67,56.67) (11.81,58.33) (11.86,60.00) (11.89,61.67) (12.57,63.33) (12.81,65.00)
    (13.03,66.67) (13.47,68.33) (13.63,70.00) (13.87,71.67) (13.89,73.33) (13.95,75.00) (14.06,76.67) (14.27,78.33)
    (14.99,80.00) (15.50,81.67) (15.54,83.33) (15.99,85.00) (15.99,86.67) (16.08,88.33) (16.31,90.00) (17.36,91.67)
    (17.45,93.33) (17.70,95.00) (18.02,96.67) (18.13,98.33) (21.81,100.00) (30.00,100.00)};
\addlegendentry{Ours / Ours (q99)}
\addplot[only marks, mark=*, mark size=1.3pt, cours, forget plot] coordinates {(21.81,100.00)};
\end{axis}
\end{tikzpicture}
\caption{Mean task progress over time on the real-robot placing bananas task, over 30 trials per condition. Each banana placed in the tray within the \qty{30}{s} limit adds half of a trial, and curves average over trials, so curves further up and to the left progress faster.}
\label{fig:zero-shot-eval}
\end{figure}

\definecolor{cpd}{HTML}{9E9E9E}
\definecolor{cdef}{HTML}{3C3C3A}
\definecolor{coursb}{HTML}{6A93D4}
\definecolor{cours}{HTML}{2A5DB0}
\begin{figure}[t]
\centering
\begin{tikzpicture}
\begin{axis}[
  width=0.62\linewidth, height=0.40\linewidth,
  xmin=0, xmax=60, ymin=0, ymax=102,
  xtick={0,10,...,60}, ytick={0,25,50,75,100},
  xlabel={Time (s)}, ylabel={Cumulative progress (\%)},
  ymajorgrids, grid style={black!8, line width=0.4pt},
  axis lines*=left, axis line style={black!55, line width=0.5pt},
  tick style={black!55}, tick align=outside, major tick length=2.5pt,
  tick label style={font=\footnotesize, text=black!70},
  label style={font=\small},
  every axis plot/.append style={const plot, line join=round, line cap=round},
  legend style={draw=none, fill=white, fill opacity=0.92, text opacity=1,
    font=\footnotesize, at={(1.04,0.5)}, anchor=west,
    row sep=1pt, inner sep=3pt},
  legend image post style={xscale=1.2},
  legend cell align=left, reverse legend, clip=false,
]
\addplot[cpd, line width=1pt, dash pattern=on 3pt off 1.8pt]
  coordinates {
    (0.00,0.00) (6.88,0.37) (7.08,0.74) (7.12,1.11) (7.20,1.48) (7.45,1.85) (7.54,2.22) (7.67,2.59)
    (7.80,2.96) (7.91,3.33) (7.92,3.70) (8.04,4.07) (8.38,4.44) (8.40,4.81) (8.43,5.19) (8.47,5.56)
    (8.52,5.93) (8.54,6.30) (8.75,6.67) (8.75,7.04) (8.81,7.41) (8.87,7.78) (8.90,8.15) (8.96,8.52)
    (9.00,8.89) (9.10,9.26) (10.08,9.63) (10.17,10.00) (10.29,10.37) (10.99,10.74) (13.66,11.11) (14.61,11.48)
    (14.67,11.85) (15.03,12.22) (15.04,12.59) (15.21,12.96) (15.72,13.33) (16.21,13.70) (16.51,14.07) (16.84,14.44)
    (17.56,14.81) (18.25,15.19) (18.47,15.56) (18.95,15.93) (19.59,16.30) (20.09,16.67) (21.19,17.04) (22.00,17.41)
    (22.38,17.78) (22.88,18.15) (23.33,18.52) (23.71,18.89) (23.88,19.26) (24.56,19.63) (24.72,20.00) (25.00,20.37)
    (25.08,20.74) (25.14,21.11) (25.40,21.48) (25.99,21.85) (26.39,22.22) (27.05,22.59) (28.52,22.96) (28.64,23.33)
    (28.99,23.70) (29.41,24.07) (29.76,24.44) (29.80,24.81) (30.46,25.19) (31.47,25.56) (32.05,25.93) (32.55,26.30)
    (33.13,26.67) (35.44,27.04) (37.59,27.41) (39.01,27.78) (39.16,28.15) (39.17,28.52) (39.56,28.89) (39.80,29.26)
    (39.85,29.63) (40.16,30.00) (40.17,30.37) (40.37,30.74) (40.60,31.11) (40.91,31.48) (41.63,31.85) (42.51,32.22)
    (43.78,32.59) (44.54,32.96) (44.63,33.33) (44.88,33.70) (45.07,34.07) (45.10,34.44) (45.20,34.81) (46.05,35.19)
    (46.70,35.56) (47.00,35.93) (47.11,36.30) (47.60,36.67) (47.87,37.04) (48.19,37.41) (48.33,37.78) (49.25,38.15)
    (50.74,38.52) (50.96,38.89) (51.13,39.26) (54.24,39.63) (54.52,40.00) (55.58,40.37) (55.84,40.74) (56.27,41.11)
    (56.40,41.48) (57.10,41.85) (58.07,42.22) (58.60,42.59) (58.96,42.96) (59.68,43.33) (60.00,43.33)};
\addlegendentry{Default / PD}
\addplot[only marks, mark=*, mark size=1.3pt, cpd, forget plot] coordinates {(59.68,43.33)};
\addplot[cdef, line width=1pt]
  coordinates {
    (0.00,0.00) (5.02,0.37) (5.53,0.74) (5.57,1.11) (5.99,1.48) (6.00,1.85) (6.00,2.22) (6.02,2.59)
    (6.06,2.96) (6.16,3.33) (6.20,3.70) (6.30,4.07) (6.33,4.44) (6.43,4.81) (6.43,5.19) (6.46,5.56)
    (6.49,5.93) (6.55,6.30) (6.56,6.67) (6.58,7.04) (6.61,7.41) (6.83,7.78) (6.87,8.15) (6.96,8.52)
    (7.10,8.89) (7.22,9.26) (7.35,9.63) (7.40,10.00) (8.26,10.37) (9.67,10.74) (10.20,11.11) (10.47,11.48)
    (11.26,11.85) (11.30,12.22) (11.38,12.59) (11.41,12.96) (11.62,13.33) (11.83,13.70) (12.21,14.07) (12.23,14.44)
    (12.33,14.81) (12.48,15.19) (12.98,15.56) (13.35,15.93) (15.13,16.30) (15.15,16.67) (16.02,17.04) (16.13,17.41)
    (17.41,17.78) (17.46,18.15) (17.53,18.52) (17.88,18.89) (18.07,19.26) (18.66,19.63) (18.70,20.00) (18.73,20.37)
    (19.37,20.74) (19.45,21.11) (19.72,21.48) (19.77,21.85) (20.42,22.22) (21.21,22.59) (21.31,22.96) (21.41,23.33)
    (21.62,23.70) (22.27,24.07) (22.38,24.44) (22.67,24.81) (22.96,25.19) (23.08,25.56) (23.35,25.93) (23.70,26.30)
    (24.05,26.67) (24.43,27.04) (24.50,27.41) (24.90,27.78) (24.90,28.15) (25.42,28.52) (25.58,28.89) (25.73,29.26)
    (26.22,29.63) (26.25,30.00) (28.44,30.37) (28.68,30.74) (28.77,31.11) (28.95,31.48) (29.14,31.85) (29.36,32.22)
    (29.39,32.59) (29.54,32.96) (29.84,33.33) (30.91,33.70) (31.28,34.07) (31.37,34.44) (31.65,34.81) (32.26,35.19)
    (32.77,35.56) (33.27,35.93) (33.34,36.30) (33.35,36.67) (33.35,37.04) (33.36,37.41) (33.82,37.78) (34.75,38.15)
    (34.77,38.52) (35.55,38.89) (35.94,39.26) (36.14,39.63) (36.24,40.00) (36.43,40.37) (36.76,40.74) (37.00,41.11)
    (37.29,41.48) (37.32,41.85) (37.97,42.22) (38.59,42.59) (39.38,42.96) (39.84,43.33) (39.85,43.70) (39.94,44.07)
    (40.11,44.44) (40.73,44.81) (41.17,45.19) (41.64,45.56) (41.77,45.93) (42.11,46.30) (42.29,46.67) (42.58,47.04)
    (43.07,47.41) (43.47,47.78) (43.75,48.15) (44.43,48.52) (44.50,48.89) (44.58,49.26) (44.77,49.63) (44.91,50.00)
    (45.02,50.37) (45.12,50.74) (45.88,51.11) (46.87,51.48) (47.09,51.85) (47.10,52.22) (48.02,52.59) (48.28,52.96)
    (48.38,53.33) (49.18,53.70) (49.52,54.07) (49.69,54.44) (49.74,54.81) (50.64,55.19) (50.78,55.56) (51.01,55.93)
    (51.31,56.30) (51.32,56.67) (52.03,57.04) (52.08,57.41) (52.11,57.78) (52.12,58.15) (52.24,58.52) (52.56,58.89)
    (52.60,59.26) (54.85,59.63) (55.06,60.00) (55.39,60.37) (55.49,60.74) (56.09,61.11) (56.17,61.48) (56.44,61.85)
    (57.17,62.22) (57.26,62.59) (57.31,62.96) (57.65,63.33) (57.79,63.70) (58.91,64.07) (59.54,64.44) (59.67,64.81)
    (59.68,65.19) (59.68,65.56) (60.00,65.56)};
\addlegendentry{Default / Ours (1x)}
\addplot[only marks, mark=*, mark size=1.3pt, cdef, forget plot] coordinates {(59.68,65.56)};
\addplot[coursb, line width=1pt]
  coordinates {
    (0.00,0.00) (4.60,0.37) (4.78,0.74) (4.80,1.11) (4.94,1.48) (4.97,1.85) (4.99,2.22) (4.99,2.59)
    (5.14,2.96) (5.18,3.33) (5.19,3.70) (5.21,4.07) (5.22,4.44) (5.28,4.81) (5.30,5.19) (5.33,5.56)
    (5.48,5.93) (5.50,6.30) (5.52,6.67) (5.55,7.04) (5.58,7.41) (5.65,7.78) (5.78,8.15) (5.91,8.52)
    (5.93,8.89) (5.96,9.26) (6.08,9.63) (6.12,10.00) (6.50,10.37) (6.60,10.74) (9.42,11.11) (9.53,11.48)
    (9.54,11.85) (9.64,12.22) (9.72,12.59) (9.79,12.96) (9.87,13.33) (9.90,13.70) (10.13,14.07) (10.44,14.44)
    (10.54,14.81) (10.79,15.19) (10.83,15.56) (10.86,15.93) (11.01,16.30) (12.50,16.67) (13.15,17.04) (13.28,17.41)
    (13.44,17.78) (13.66,18.15) (14.16,18.52) (14.53,18.89) (15.02,19.26) (15.03,19.63) (15.27,20.00) (15.31,20.37)
    (15.36,20.74) (15.46,21.11) (15.55,21.48) (15.57,21.85) (15.74,22.22) (16.45,22.59) (17.05,22.96) (17.13,23.33)
    (17.40,23.70) (17.43,24.07) (17.88,24.44) (18.16,24.81) (18.59,25.19) (18.69,25.56) (18.79,25.93) (18.84,26.30)
    (19.33,26.67) (20.00,27.04) (20.10,27.41) (20.16,27.78) (20.35,28.15) (21.32,28.52) (21.38,28.89) (21.41,29.26)
    (21.57,29.63) (21.58,30.00) (21.70,30.37) (21.74,30.74) (22.27,31.11) (22.33,31.48) (22.77,31.85) (22.90,32.22)
    (23.37,32.59) (23.64,32.96) (23.66,33.33) (23.79,33.70) (23.95,34.07) (24.31,34.44) (24.32,34.81) (24.75,35.19)
    (24.93,35.56) (25.15,35.93) (25.22,36.30) (25.47,36.67) (25.64,37.04) (25.73,37.41) (25.73,37.78) (26.00,38.15)
    (26.10,38.52) (27.06,38.89) (27.13,39.26) (27.98,39.63) (28.40,40.00) (28.94,40.37) (29.07,40.74) (29.35,41.11)
    (29.57,41.48) (29.59,41.85) (29.98,42.22) (30.02,42.59) (30.11,42.96) (30.27,43.33) (30.73,43.70) (30.78,44.07)
    (31.30,44.44) (31.39,44.81) (31.44,45.19) (31.56,45.56) (31.85,45.93) (31.93,46.30) (32.00,46.67) (32.04,47.04)
    (32.81,47.41) (33.07,47.78) (33.36,48.15) (34.86,48.52) (35.36,48.89) (35.40,49.26) (35.61,49.63) (35.80,50.00)
    (35.95,50.37) (35.98,50.74) (36.14,51.11) (36.19,51.48) (36.33,51.85) (36.92,52.22) (36.94,52.59) (36.97,52.96)
    (37.18,53.33) (37.61,53.70) (37.65,54.07) (37.72,54.44) (37.82,54.81) (38.35,55.19) (38.42,55.56) (38.45,55.93)
    (39.43,56.30) (39.53,56.67) (40.05,57.04) (40.73,57.41) (40.95,57.78) (41.18,58.15) (41.30,58.52) (41.90,58.89)
    (41.95,59.26) (42.07,59.63) (42.13,60.00) (42.33,60.37) (42.61,60.74) (42.65,61.11) (42.81,61.48) (43.02,61.85)
    (43.47,62.22) (43.62,62.59) (43.74,62.96) (43.83,63.33) (43.97,63.70) (44.19,64.07) (44.37,64.44) (44.77,64.81)
    (44.83,65.19) (44.94,65.56) (45.78,65.93) (46.55,66.30) (46.72,66.67) (47.03,67.04) (47.42,67.41) (47.48,67.78)
    (48.15,68.15) (48.32,68.52) (48.55,68.89) (48.80,69.26) (48.82,69.63) (49.24,70.00) (49.56,70.37) (49.56,70.74)
    (49.93,71.11) (50.39,71.48) (50.82,71.85) (51.11,72.22) (51.61,72.59) (51.87,72.96) (51.93,73.33) (52.16,73.70)
    (52.23,74.07) (52.53,74.44) (52.67,74.81) (52.99,75.19) (53.80,75.56) (54.53,75.93) (55.52,76.30) (55.83,76.67)
    (56.18,77.04) (56.32,77.41) (56.63,77.78) (56.84,78.15) (57.08,78.52) (57.24,78.89) (57.41,79.26) (57.57,79.63)
    (57.59,80.00) (57.71,80.37) (58.77,80.74) (58.88,81.11) (59.59,81.48) (59.61,81.85) (59.80,82.22) (59.87,82.59)
    (60.00,82.59)};
\addlegendentry{Ours / Ours (1x)}
\addplot[only marks, mark=*, mark size=1.3pt, coursb, forget plot] coordinates {(59.87,82.59)};
\addplot[cours, line width=1.5pt]
  coordinates {
    (0.00,0.00) (3.88,0.37) (4.03,0.74) (4.08,1.11) (4.10,1.48) (4.14,1.85) (4.17,2.22) (4.28,2.59)
    (4.29,2.96) (4.31,3.33) (4.33,3.70) (4.34,4.07) (4.36,4.44) (4.38,4.81) (4.40,5.19) (4.44,5.56)
    (4.47,5.93) (4.49,6.30) (4.49,6.67) (4.51,7.04) (4.52,7.41) (4.69,7.78) (4.71,8.15) (4.92,8.52)
    (5.13,8.89) (5.15,9.26) (5.16,9.63) (6.44,10.00) (7.29,10.37) (7.89,10.74) (7.98,11.11) (8.05,11.48)
    (8.06,11.85) (8.12,12.22) (8.36,12.59) (8.42,12.96) (8.48,13.33) (8.51,13.70) (8.53,14.07) (8.54,14.44)
    (8.55,14.81) (8.66,15.19) (8.70,15.56) (8.74,15.93) (8.88,16.30) (8.98,16.67) (9.04,17.04) (9.26,17.41)
    (9.32,17.78) (9.53,18.15) (10.40,18.52) (10.75,18.89) (10.76,19.26) (10.91,19.63) (12.31,20.00) (12.43,20.37)
    (12.46,20.74) (12.89,21.11) (12.96,21.48) (13.07,21.85) (13.55,22.22) (13.58,22.59) (13.64,22.96) (13.68,23.33)
    (13.77,23.70) (13.78,24.07) (13.82,24.44) (13.96,24.81) (14.68,25.19) (15.09,25.56) (15.30,25.93) (15.85,26.30)
    (15.97,26.67) (16.09,27.04) (16.42,27.41) (16.42,27.78) (16.59,28.15) (16.74,28.52) (17.11,28.89) (17.17,29.26)
    (17.20,29.63) (17.40,30.00) (18.00,30.37) (18.19,30.74) (18.25,31.11) (18.53,31.48) (18.56,31.85) (18.68,32.22)
    (18.69,32.59) (18.71,32.96) (18.73,33.33) (18.76,33.70) (18.79,34.07) (18.81,34.44) (18.82,34.81) (18.83,35.19)
    (19.59,35.56) (19.77,35.93) (20.48,36.30) (20.53,36.67) (20.70,37.04) (20.78,37.41) (21.29,37.78) (21.30,38.15)
    (21.83,38.52) (22.02,38.89) (22.17,39.26) (22.56,39.63) (22.62,40.00) (22.93,40.37) (23.00,40.74) (23.15,41.11)
    (23.35,41.48) (23.42,41.85) (23.74,42.22) (23.79,42.59) (23.84,42.96) (24.02,43.33) (24.08,43.70) (24.16,44.07)
    (24.37,44.44) (24.53,44.81) (24.78,45.19) (25.02,45.56) (25.21,45.93) (25.62,46.30) (25.71,46.67) (25.84,47.04)
    (25.90,47.41) (25.94,47.78) (26.01,48.15) (27.13,48.52) (27.35,48.89) (27.52,49.26) (27.64,49.63) (27.74,50.00)
    (27.93,50.37) (27.95,50.74) (28.46,51.11) (28.58,51.48) (28.86,51.85) (28.86,52.22) (29.63,52.59) (29.84,52.96)
    (30.00,53.33) (30.13,53.70) (30.24,54.07) (30.48,54.44) (30.70,54.81) (31.43,55.19) (31.64,55.56) (31.91,55.93)
    (32.06,56.30) (32.30,56.67) (32.32,57.04) (32.39,57.41) (32.80,57.78) (33.04,58.15) (33.28,58.52) (33.65,58.89)
    (33.84,59.26) (34.07,59.63) (34.52,60.00) (34.84,60.37) (35.26,60.74) (35.42,61.11) (35.56,61.48) (35.86,61.85)
    (35.91,62.22) (36.42,62.59) (36.81,62.96) (37.18,63.33) (37.49,63.70) (37.60,64.07) (37.60,64.44) (38.08,64.81)
    (38.09,65.19) (38.13,65.56) (38.45,65.93) (38.46,66.30) (38.51,66.67) (38.54,67.04) (38.73,67.41) (39.04,67.78)
    (39.81,68.15) (40.08,68.52) (40.10,68.89) (40.47,69.26) (40.74,69.63) (42.01,70.00) (42.33,70.37) (42.79,70.74)
    (42.88,71.11) (42.95,71.48) (43.05,71.85) (43.15,72.22) (43.26,72.59) (43.88,72.96) (44.94,73.33) (44.98,73.70)
    (45.07,74.07) (45.16,74.44) (45.18,74.81) (45.39,75.19) (45.82,75.56) (46.22,75.93) (46.23,76.30) (46.29,76.67)
    (46.39,77.04) (46.65,77.41) (47.38,77.78) (47.83,78.15) (48.42,78.52) (48.59,78.89) (48.84,79.26) (49.37,79.63)
    (49.56,80.00) (49.64,80.37) (49.77,80.74) (50.57,81.11) (50.77,81.48) (50.83,81.85) (50.97,82.22) (51.24,82.59)
    (52.03,82.96) (52.13,83.33) (52.90,83.70) (52.95,84.07) (52.95,84.44) (53.39,84.81) (53.69,85.19) (53.88,85.56)
    (53.89,85.93) (54.59,86.30) (54.76,86.67) (54.85,87.04) (55.24,87.41) (55.52,87.78) (55.53,88.15) (56.03,88.52)
    (56.83,88.89) (57.08,89.26) (57.09,89.63) (57.27,90.00) (57.78,90.37) (58.39,90.74) (58.80,91.11) (59.31,91.48)
    (59.76,91.85) (60.00,91.85)};
\addlegendentry{Ours / Ours (q99)}
\addplot[only marks, mark=*, mark size=1.3pt, cours, forget plot] coordinates {(59.76,91.85)};
\end{axis}
\end{tikzpicture}
\caption{Mean task progress over time on the real-robot placing cubes task, over 30 trials per condition. Each cube placed in the bin within the \qty{60}{s} limit adds one ninth of a trial, and curves average over trials, so curves further up and to the left progress faster.}
\label{fig:zero-shot-cube}
\end{figure}

\end{document}